\documentclass[a4paper]{article}
\usepackage{graphicx}
\usepackage{mystylefile}
\usepackage{subfigure}
\usepackage{times}
\usepackage{color}
\usepackage{multirow}
\usepackage[natbibapa]{apacite}
\usepackage{epstopdf}
\usepackage[np,autolanguage]{numprint}

\pdfoutput=1

\newcommand{\M}{\mathbf{M}} 
\newcommand{\m}{\vector{m}}
\newcommand{\mhat}{\widehat{\vector{m}}}
\newcommand{\g}{\vector{g}}
\newcommand{\h}{\vector{h}}

 \newcommand{\U}{\mathbf{U}}
 
\newcommand{\Lam}{\mathbf{\Lambda}} \newcommand{\The}{\mathbf{\Theta}}
\newcommand{\bfOme}{\mathbf{\Omega}} 
\DeclareMathOperator*{\argmin}{arg\,min}

\graphicspath{{./Figs/}}

\title{Simultaneous Estimation of Non-Gaussian Components and their
Correlation Structure}
\author{Hiroaki Sasaki \\ hsasaki@is.naist.jp \\ Graduate School
of Information Science\\ Nara Institute of Science and Technology, Japan 
\\\vspace{-1mm} \\
Michael U. Gutmann \\ michael.gutmann@ed.ac.uk \\ 
School of Informatics\\ University of Edinburgh, United Kingdom
\\\vspace{-1mm} \\
Hayaru Shouno \\ shouno@uec.ac.jp\\ Graduate School of Informatics and
Engineering\\ The University of Electro-Communications, Japan
\\\vspace{-1mm} \\
Aapo Hyv{\"a}rinen \\ a.hyvarinen@ucl.ac.uk \\ Department of Computer
Science \\ Helsinki Institute for Information Technology HIIT\\
University of Helsinki, Finland\\ Gatsby Computational Neuroscience
Unit\\ University College London, United Kingdom}
\date{}
\begin{document} 
 \maketitle

 \begin{abstract}%
  The statistical dependencies which independent component analysis
  (ICA) cannot remove often provide rich information beyond the linear
  independent components. It would thus be very useful to estimate the
  dependency structure from data. While such models have been proposed,
  they usually concentrated on higher-order correlations such as energy
  (square) correlations. Yet, linear correlations are a most fundamental
  and informative form of dependency in many real data sets. Linear
  correlations are usually completely removed by ICA and related
  methods, so they can only be analyzed by developing new methods which
  explicitly allow for linearly correlated components. In this paper, we
  propose a probabilistic model of linear non-Gaussian components which
  are allowed to have both linear and energy correlations. The precision
  matrix of the linear components is assumed to be randomly generated by
  a higher-order process and explicitly parametrized by a parameter
  matrix. The estimation of the parameter matrix is shown to be
  particularly simple because using score
  matching~\citep{hyvarinen2005estimation}, the objective function is a
  quadratic form. Using simulations with artificial data, we demonstrate
  that the proposed method improves identifiability of non-Gaussian
  components by simultaneously learning their correlation
  structure. Applications on simulated complex cells with natural image
  input, as well as spectrograms of natural audio data show that the
  method finds new kinds of dependencies between the components.
 \end{abstract}

\section{Introduction}
\label{sec:int}
Estimating latent non-Gaussian components is important in modern
statistical data analysis and machine learning. A well-known method for
that purpose is independent component analysis
(ICA)~\citep{comon1994independent,hyvarinen2000independent}, whose goal
is to identify non-Gaussian components as statistically independent as
possible. ICA has been applied in a wide range of fields such as brain
imaging analysis~\citep{vigario2000independent}, image processing
\citep{Hyvarinen2009}, pattern recognition~\citep{bartlett2002face}, or
causal analysis~\citep{shimizu2006linear}.

The components estimated by ICA, however, are often not independent at
all. For natural images, for instance, the estimated components may have
variance dependencies, that is, the squares of the components may be
correlated, and the same holds also for the wavelet
coefficients~\citep{simoncelli1999modeling,hyvarinen2000emergence,karklin2005hierarchical}.
Inspired by this fact, extensions of ICA have been developed which take
dependencies between the components into account: Independent subspace
analysis (ISA), which combines the techniques of multidimensional
ICA~\citep{cardoso1998multidimensional,theis2005blind} and the principle
of invariant-feature
subspaces~\citep{kohonen1995self,kohonen1996emergence}, divides the
components into pre-defined groups where the components in each group
have variance dependencies~\citep{hyvarinen2000emergence}. When applied
to natural images, ISA produces a phase invariant pooling of the
non-Gaussian components. Methods for learning topographic representation
assume that nearby components have statistical dependencies, while
far-away components are as statistically independent as
possible~\citep{hyvarinen2001topographic,mairal2011convex,sasaki2013correlated}.
The dependencies are used to order the components and to arrange them on
a grid, which provides a convenient visualization of properties of the
data.  Thus, the statistical dependencies which ICA cannot remove often
contain rich information.  However, a limitation of the work cited above
is that it assumes pre-fixed dependency structures, which can be
problematic because specifying a wrong dependency structure can hamper
the identifiability of the non-Gaussian
components~\citep{sasaki2013correlated}.

\begin{figure}[!t]
  \centering
  \subfigure{\includegraphics[width=0.32\textwidth]{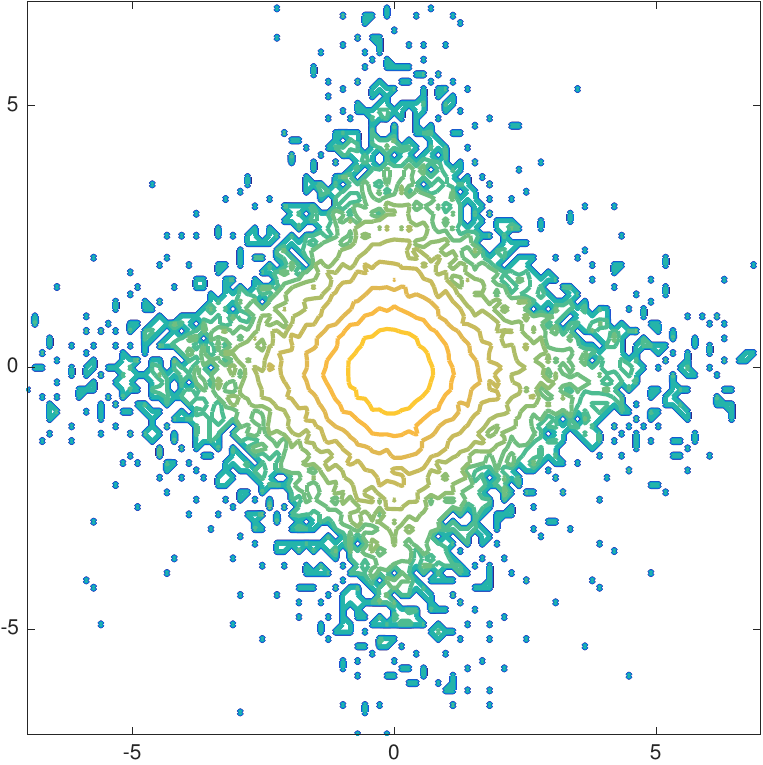}}
  \subfigure{\includegraphics[width=0.32\textwidth]{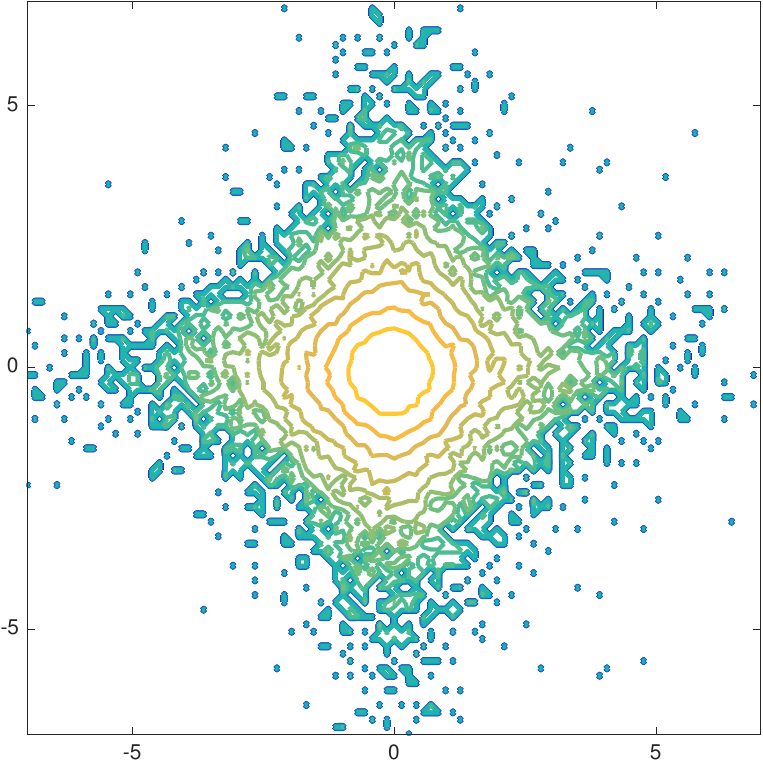}}
  \subfigure{\includegraphics[width=0.32\textwidth]{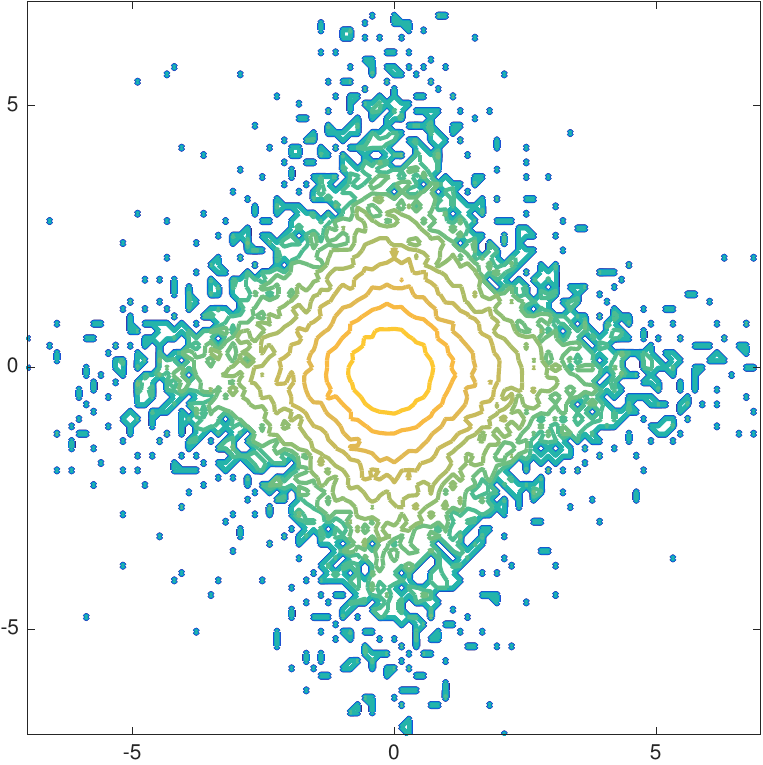}}\\
  \setcounter{subfigure}{0}
  \subfigure[Source]{\includegraphics[width=0.32\textwidth]{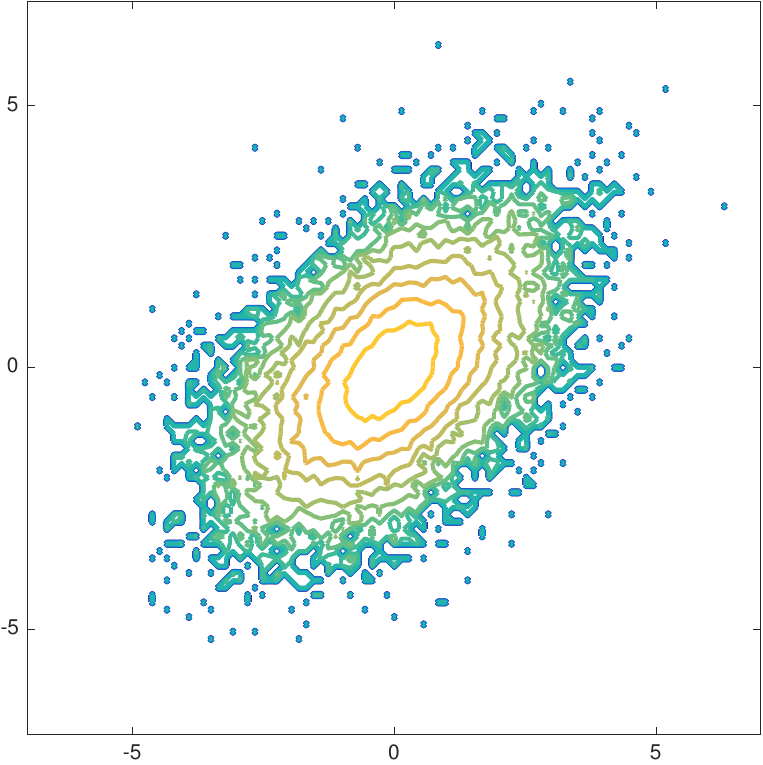}}
  \subfigure[ICA]{\includegraphics[width=0.32\textwidth]{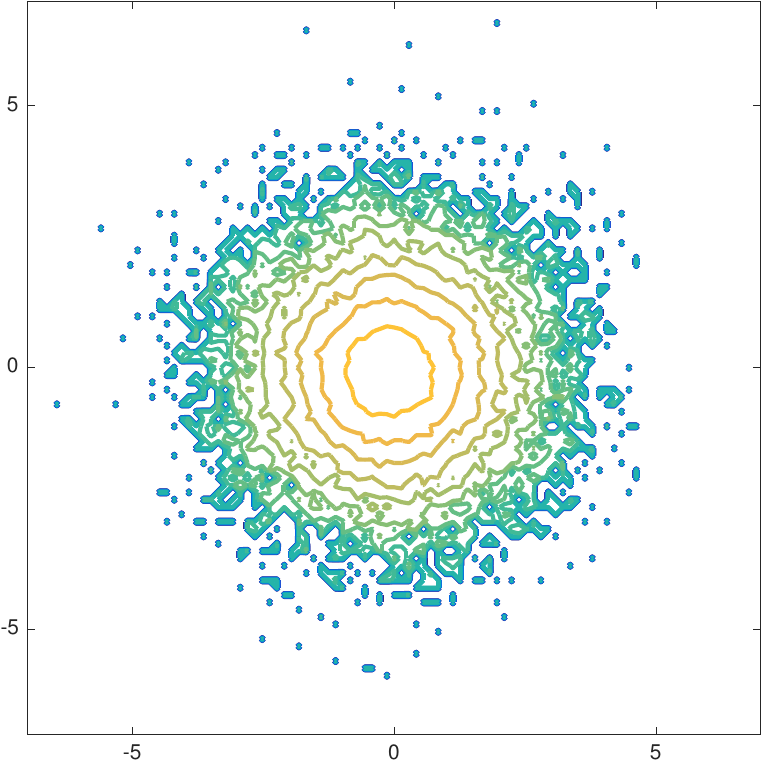}}
  \subfigure[Proposed
  Method]{\includegraphics[width=0.32\textwidth]{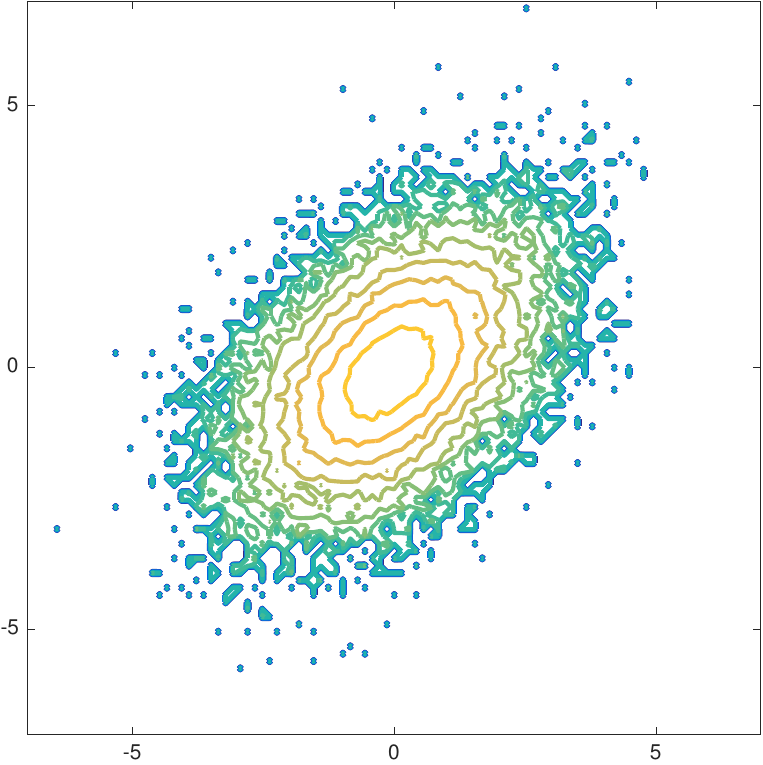}}
  \caption{\label{fig:Example} The logarithms of the non-parametric
  estimates of the two-dimensional probability density functions (a) for
  artificially generated source components, and (b,c) for estimated
  source components by ICA (maximum likelihood estimation without the
  decorrelation constraint as in~\eqref{eqn:W0}) and the proposed
  method, respectively. The first row is for independent non-Gaussian
  components, while the second one is for linearly correlated ones. The
  correlation coefficients in the second row are (a) 0.499, (b) 0.01 and
  (c) 0.472, respectively.}
 \end{figure}

This limitation can be removed by estimating dependency structures
themselves from data. Two-layer models are suitable for this purpose, if
they further incorporate some higher-order process for non-Gaussian
components: \citet{karklin2005hierarchical} proposed a method to
estimate a two-layer model whose first layer consists of ICA-like
components and whose second layer represents density
components. \citet{osindero2006topographic} proposed a two-layer
topographic model with non-Gaussian components in the first layer and
weighted connections among them in the second layer. A related two-layer
model was proposed by~\citet{koster2010two}. However, most of the models
focus on higher-order dependencies only, and ignore linear correlations,
usually even assuming that they are zero. But linear correlations
between non-Gaussian components can be observed in real data for a
couple of models~\citep{gomez2008measuring,coen2012cortical}. It is thus
probable that the underlying components which are forced to be linearly
uncorrelated in the estimation would actually be correlated. For
instance, Figure~\ref{fig:Example} illustrates that the components
estimated by ICA are linearly uncorrelated even when the underlying
source components are correlated. Therefore, it would be meaningless to
analyze linear correlations in components estimated by ordinary ICA
methods; it is necessary to incorporate the linear correlations in the
very definition of the model.  Along these lines, the estimation of
topographic representations has been recently improved by taking into
account both linear and variance
correlations~\citep{sasaki2013correlated}. Linear and higher-order
dependencies between the components have also been exploited in order to
find correspondences between features in multiple data sets
\citep{Gutmann2011a,Campi2013,Gutmann2014}.
\setcounter{subfigure}{0}

 In this paper, we propose a novel method to estimate latent
 non-Gaussian components and their dependency structure
 simultaneously. The dependency structure includes both linear and
 higher-order correlations, and is parametrized by a single matrix. The
 off-diagonal elements of this dependency matrix represent the
 conditional dependencies much like the precision matrix does for
 Gaussian Markov random fields. More generally, the dependency matrix
 defines a distance matrix which can be used for visualization via an
 undirected graph, multidimensional scaling, or some other suitable
 technique. The proposed method can be interpreted as a generalization
 of ICA and correlated topographic analysis
 (CTA)~\citep{sasaki2013correlated} where the dependency structures are
 assumed to be known. To develop the method, we begin with a new
 generative model for sources, which generalizes previous models
 of~\citet{hyvarinen2001topographic,karklin2005hierarchical,osindero2006topographic,koster2010two}
 by capturing linear correlations between the components. The previous
 models generate non-Gaussian components without linear, but with
 higher-order correlations~\citep[Section~9.3]{Hyvarinen2009}. Divisive
 normalization
 models~\citep{heeger1992normalization,schwartz2001natural,balle2015density}
 are also closely related, and focus on higher-order correlations.

 Estimating two-layer models or Markov random fields for non-Gaussian
components is often difficult because sophisticated parametric models
generally have an intractable partition function so that the standard
maximum likelihood estimation cannot be applied. To cope with this
problem, several estimation methods have been proposed, such as
contrastive divergence~\citep{hinton2002training}, score
matching~\citep{hyvarinen2005estimation}, or noise contrastive
estimation~\citep{Gutmann2012a} and its
extensions~\citep{Pihlaja2010,Gutmann2011b} (see \citet{Gutmann2013b}
for an introductory paper). Score matching has a particularly useful
property for the proposed model: The objective function for the
estimation of the dependency parameters takes a simple quadratic form,
and can be optimized by standard quadratic programming. Due to this
computational simplification, here we use score matching, and
empirically show that our method estimates the dependency structure and
improves identifiability of the non-Gaussian components.

The paper is organized as follows: In Section~\ref{sec:esti}, we begin
with a novel probabilistic generative model for conditional precision
matrices. Based on the generative model, we derive an approximation of
the marginal density for non-Gaussian components where the dependency
structure is explicitly parametrized like a precision
matrix. Section~\ref{sec:algo} deals with estimation of the model,
presenting a powerful algorithm for identifying the non-Gaussian
components and their dependency structure.  In Section~\ref{sec:ide}, we
perform numerical experiments and compare the proposed method with
existing methods on artificial data. Section~\ref{sec:real} demonstrates
the applicability of the method to real data. Connections to past work
and extensions of the proposed method are discussed in
Section~\ref{sec:disc}. Section~\ref{sec:conc} concludes this paper. A
preliminary version of this paper was presented at
AISTATS~2014~\citep{Sasaki2014esti}.
\section{Generative Model with Dependent Non-Gaussian Components}
\label{sec:esti}
Here, we introduce a novel generative model with dependent non-Gaussian
components. The probability density function (pdf) of the components and
the data is shown to be only implicitly defined via an intractable
integral. We derive an approximation of the pdf where the dependency
structure of the components is explicitly parametrized, and demonstrate
the validity of the approximation using both analytical arguments and
simulations.
\subsection{The Generative Model}
\label{ssec:model}
As in previous work related to ICA~\citep{hyvarinen2000independent}, we
assume a linear mixing model for the data,
\begin{align}
 \vector{x}&=s_1\vector{a}_1+s_2\vector{a}_2+\dots+s_d\vector{a}_d
 =\A\vector{s}, \label{eqn:datamodel}
\end{align}
where $\vector{x}=(x_1,x_2,\dots,x_d)^{\top}$ denotes the
$d$-dimensional data vector,
$\A=(\vector{a}_1,\vector{a}_2,\dots,\vector{a}_d)$ is the $d$ by $d$
mixing matrix formed by the basis vectors $\vector{a}_i$, and
$\vector{s}=(s_1,s_2,\dots,s_d)^{\top}$ is the $d$-dimensional vector
consisting of the latent non-Gaussian components (the sources). The
non-Gaussianity assumption about $\vector{s}$ is fundamental for the
identification of the mixing model~\citep{comon1994independent}.

We next construct a model for the components $s_i$ which allows them to
be statistically dependent, in contrast to ICA. We assume that
$\vector{s}$ is generated from a Gaussian distribution with precision
matrix $\Lam$ whose elements $\lambda_{ij}$ of $\Lam$ are random
variables themselves, generated before $\vector{s}$ in a higher level of
hierarchy as
\begin{align}
 \lambda_{ij}=\left\{
 \begin{array}{cc}
  -u_{ij}, & i\neq j,\\
  \sum_{k=1}^d u_{ik}, & i=j.
 \end{array}
 \right.\label{eqn:lambda}
\end{align}
The $u_{ij}$ are independent non-negative random variables, and because
precision matrices are symmetric, we set $u_{ij}=u_{ji}$. To ensure
invertability, we also require that $u_{ii}>0$. Nonzero $u_{ij}$ produce
positive correlations between $s_i$ and $s_j$ (given the remaining
variables), and larger values of $u_{ij}$ result in more strongly
correlated variables. The motivation for the definition of the diagonal
elements $\lambda_{ii}$ is that it guarantees realizations of $\Lam$
which are invertible and positive definite ($\Lam$ is symmetric and
strictly diagonally dominant, from which the stated properties follow
(see Theorem 6.1.10 in~\citet{horn1985matrix} and
Appendix~\ref{app:general})). Readers familiar with graph theory will
recognize that $\Lam$ equals the Laplacian matrix of a weighted graph
defined by the matrix $\U$ with elements
$u_{ij}$~\citep{bollobas1998modern}.

Another property of the model is that the components $s_i$ are
super-Gaussian: \citet{hyvarinen2001topographic} showed that the
marginal pdfs of the Gaussian variables with random variances have
heavier tails than a Gaussian pdf. Furthermore, since the conditional
variances are dependent on each other in model (\ref{eqn:lambda}),
higher-order correlations are likely to
exist~\citep{hyvarinen2001topographic,sasaki2013correlated}.

We thus assume that the conditional pdf of $\vector{s}$ given
$\Lam$ equals $p(\vector{s}|\Lam)$,
\begin{align}
  p(\vector{s}|\Lam) &=\frac{|\Lam|^{1/2}}{(2\pi)^{d/2}}
  \exp\left(-\frac{1}{2}\vector{s}^{\top} \Lam\vector{s}\right)\\ 
  &=\frac{|\Lam|^{1/2}}{(2\pi)^{d/2}}
  \exp\left(-\frac{1}{2}\sum_{i=1}^d\left\{\lambda_{ii}s_i^2
 +\sum_{j\neq i} \lambda_{ij}s_is_j\right\}\right)
 \label{eqn:condPDFtmp}, 
\end{align}
which we can write in terms of the $u_{ij}$ as $p(\vector{s}|\U)$,
\begin{align}
 p(\vector{s}|\U)&=\frac{|\Lam|^{1/2}}{(2\pi)^{d/2}}
 \exp\left(-\frac{1}{2}\sum_{i=1}^d \left\{s_i^2u_{ii} +\sum_{j> i}
 \left(s_i-s_j\right)^2u_{ij} \right\}\right),  \label{eqn:condPDF} 
\end{align}
as proved in Appendix~\ref{app:calc}. While not explicitly visible
from the notation, the determinant $|\Lam|$ is a function of the
$u_{ij}$.

The specification of the distribution of the $u_{ij}$ completes the
model, but this is a complex issue which we postpone to the next
subsection. Denoting the pdf of the $u_{ij}$ generally by $p_u$, the pdf
of the sources equals $p_s(\vector{s})$,
\begin{equation}
p_s(\vector{s}) = \int_{0}^{\infty} p(\vector{s}|\U)p_u(\U) d\U.
\label{eqn:intgen}
\end{equation}
and the pdf of $\vector{x}$ follows from the standard formula for linear
transformations of random variables,
\begin{equation}
  p_x(\vector{x}) = p_s(\W \vector{x}) |\W|,\label{eqn:px}
\end{equation}
with $\W = \A^{-1}$. However, since the determinant $|\Lam|$ in
$p(\vector{s}|\U)$ depends on $\U$, solving the multi-dimensional
integral in (\ref{eqn:intgen}) is practically impossible for any choice
of $p_u$. While the integral can be estimated using Monte Carlo methods,
it could be time-consuming. Therefore, we consider next an analytical
approximation of the determinant which allows us to find an
approximation of $p_s$ that holds qualitatively for any $p_u$. Once we
have an approximation, $\tilde{p}_s$ say, we can use (\ref{eqn:px}) to
obtain a tractable approximation $\tilde{p}_x$ of the pdf of
$\vector{x}$,
\begin{equation}
 \tilde{p}_x(\vector{x}) = \tilde{p}_s(\W \vector{x})|\W|.\label{eqn:tildepx}
\end{equation}
\subsection{Approximating the Density of the Dependent Non-Gaussian Components}
\label{ssec:derive}
In order to derive an approximation $\tilde{p}_s$ of $p_s$, we
approximate the determinant of $\Lam$ via a product over the $u_{ii}$,
\begin{align}
 |\Lam| &\approx \prod_{i=1}^d u_{ii} .\label{eqn:approx}
\end{align}
This is the only approximation which we need to obtain the tractable
$\tilde{p}_s$ below. Another meaning of this approximation is to give a
lower bound of $p(\vector{s}|\U)$, and consequently $\tilde{p}_s$ is a
lower-bound of $p_s$, which is proved by using the Ostrowski's
inequality in Appendix~\ref{app:lower}. In other words, $\tilde{p}_s$ is
an unnormalized model which is defined up to a multiplicative factor not
depending on $\vector{s}$. This is not an insurmountable problem but
needs to be taken into account when performing the estimation
\citep[see, for example, ][]{Gutmann2013b}.

Inserting the approximation (\ref{eqn:approx}) and the independence
assumption of the $u_{ij}$ into (\ref{eqn:intgen}) yields the following
approximation of $p_s(\vector{s})$:
\begin{align}
 p_s(\vector{s}) \approx & \tilde{p}_s(\vector{s})\nonumber\\ \propto &
 \int_0^{\infty}\left[\prod_{i=1}^d \sqrt{u_{ii}}\right]
 \exp\left(-\frac{1}{2}\sum_{i=1}^d \left\{s_i^2u_{ii} +\sum_{j> i}
 \left(s_i-s_j\right)^2u_{ij} \right\}\right) p_u(\U) d\U\\ \propto&
 \int_0^{\infty} \left[\prod_{i=1}^d \sqrt{u_{ii}}\right]
 \left[\prod_{i=1}^d \exp\left(-\frac{1}{2} s_i^2u_{ii}\right)
 \prod_{j>i} \exp\left(-\frac{1}{2}(s_i-s_j)^2u_{ij}\right)\right]
 \times \nonumber\\ & \left[\prod_{i=1}^d \prod_{j>i}
 p_{ij}(u_{ij})\right] d\U,
\end{align}
where the product over the $p_{ij}$ in the last line is the pdf $p_u$ of the
$u_{ij}$ due to their independence. The expression for $\tilde{p}_s$ can be simplified by grouping together terms
featuring $u_{ii}$ and $u_{ij}$ only,
\begin{align}
  \tilde{p}_s(\vector{s}) \propto&  \prod_{i=1}^d \left[\int_0^{\infty}
    \sqrt{u_{ii}} \exp\left(-\frac{1}{2} s_i^2u_{ii}
    \right) p_{ii}(u_{ii}) d u_{ii} \right] \times \nonumber \\
 &  \prod_{j>i} \left[\int_0^{\infty} \exp\left (-\frac{1}{2}
    \left(s_i-s_j\right)^2u_{ij}\right)p_{ij}(u_{ij}) d
  u_{ij}\right]\\
 \propto&  \prod_{i=1}^d g_{ii}(s_i^2) \prod_{j>i} g_{ij}( (s_i-s_j)^2 ),
\end{align}
where we have introduced the non-negative functions $g_{ii}(v)$ and
$g_{ij}(v)$, defined for $v \ge 0$,
\begin{align} 
 g_{ii}(v) &\propto \int_0^{\infty} \sqrt{u_{ii}}\exp\left(-\frac{v}{2} u_{ii} \right) p_{ii}(u_{ii}) d u_{ii}, \\ g_{ij}(v) &\propto \int_0^{\infty} \exp\left(-\frac{v}{2}
    u_{ij}\right)p_{ij}(u_{ij}) d u_{ij} \;\text{ for }\; i\neq j.\label{eqn:g_def}
\end{align}
The proportionality sign is used because $\tilde{p}_s$ is only defined
up to the partition function. The approximation of the determinant thus
allowed us to transform the multidimensional integral in
(\ref{eqn:intgen}) into a product of functions which are defined via
one-dimensional integrals. The one-dimensional integrals can be easily
solved numerically for arbitrary $p_{ij}$, or also analytically for
particular choices of them. We also note that the $g_{ij}$ are related
to the Laplace transform of the $p_{ij}$.

Different pdfs $p_{ij}$ yield different functions $g_{ij}$. But,
Appendix \ref{app:logconvex} shows that unless $g_{ij}(v)$ is a
constant, the different $\log g_{ij}(v)$ are monotonically decreasing
convex functions for any choice of $p_{ij}$. We thus focus on the
following particular class of functions
\begin{equation}
  \log g_{ij}(v) = -m_{ij} \sqrt{v} + \mathrm{const}. \label{eqn:chosen_g}
\end{equation}
The $m_{ij}$ are free parameters which will be estimated from the
data. They are symmetric, $m_{ij} = m_{ji}$. Further, we require that
$m_{ii}>0$ so that $\tilde{p}_s$ depends on all $s_i$. For $m_{ij}$,
$i\neq j$, we only require non-negativity: If $m_{ij}=0$, then
$g_{ij}((s_i-s_j)^2) = \mathrm{const}$ which happens when the variable
$u_{ij}$ is deterministically zero.

The particular choice (\ref{eqn:chosen_g}) is motivated by its
simplicity, but we show in Appendix~\ref{app:integral} that it
corresponds to choosing an inverse-Gamma distribution for the $u_{ij}$,
\begin{equation}
 p_{ij}(u_{ij}) =
  \left\{
   \begin{array}{cc}
    \frac{m_{ii}^2}{2}u_{ii}^{-2}
     \exp\left(-\frac{m^2_{ii}}{2u_{ii}}\right), & i = j,\\    
    \frac{m_{ij}}{\sqrt{2\pi}} u_{ij}^{-3/2}
     \exp\left(-\frac{m^2_{ij}}{2u_{ij}}\right), & i \neq j.     
   \end{array}
  \right. \label{eqn:pij_def}
\end{equation}
The parameters $m_{ij}$ determine the mode of the $p_{ij}$ (i.e. the
point at which $p_{ij}$ is maximum): The mode is $m_{ii}^2/2$ for $i=j$
and $2m_{ij}^2/3$ otherwise.

Denoting by $\M$ the matrix formed by the $m_{ij}$, and its
upper-triangular part by $\m$, $\m = (m_{11}, \ldots m_{1d}, m_{22},
\ldots m_{2d},m_{33},\ldots,m_{3d},\ldots,m_{dd})$, we obtain the
approximation $\tilde{p}_s(\vector{s}) = \tilde{p}_s(\vector{s};\m)$,
\begin{align}
 \tilde{p}_s(\vector{s};\m) \propto \prod_{i=1}^d
  \exp\left(-m_{ii}|s_i| -\sum_{j>i}m_{ij} |s_i-s_j|\right),
  \label{eqn:PDF}
\end{align}
which we will use in the following sections.  The terms $|s_i|$ in
(\ref{eqn:PDF}) are related to modelling the $s_i$ as super-Gaussian,
and the terms $|s_i-s_j|$ capture statistical dependencies between the
components. The dependencies can be read out from the dependency matrix
$\M$: If $m_{ij}=0$ for some $j\neq i$, $s_i$ is independent from the
$s_j$ conditioned on the other variables. Furthermore, larger $m_{ij}$
imply stronger conditional (positive) dependencies between $s_i$ and
$s_j$.

The model (\ref{eqn:PDF}) generalizes pdfs used in previous work in the
following ways:
\begin{enumerate}
 \item $\tilde{p}$ approaches the Laplacian factorizable pdf when
       $m_{ij}\rightarrow 0$ for all $j\neq i$. Laplacian factorizable
       pdfs are often assumed for super-Gaussian components in ICA.
      
 \item $\tilde{p}$ approaches the topographic pdf in
       CTA~\citep{sasaki2013correlated} when $m_{ii}\rightarrow 1$,
       $m_{i,i+1}\rightarrow 1$ for all $i$ and $m_{ij}\rightarrow 0$
       otherwise.  The topographic pdf was derived using a different
       generative model and resorting to two quite heuristic
       approximations. This is contrast to this paper where a single
       relatively well-justified approximation was used. The new
       derivation is not only more elegant, but it allows for further
       extensions of the model as well, which is done in
       Appendix~\ref{app:general}.

 \item $\tilde{p}$ was heuristically proposed in our preliminary
       conference paper \citep{Sasaki2014esti} as a simple extension of
       CTA. In this paper, on the other hand, we derived $\tilde{p}$
       from a novel generative model for random precision matrices.
\end{enumerate}
\subsection{Numerical Validation of the Approximation}
\setcounter{subfigure}{0}
 \begin{figure}[!t]
  \centering
  \subfigure[Model]{\includegraphics[width=0.32\textwidth]{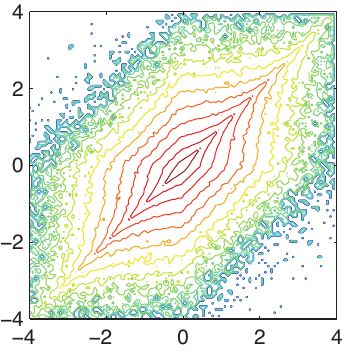}}
  \subfigure[Approximation]{\includegraphics[width=0.32\textwidth]{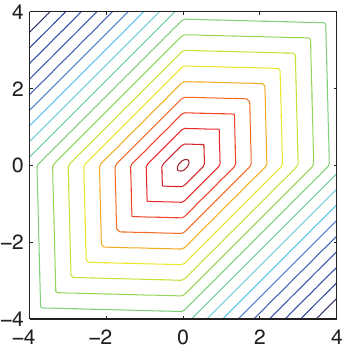}}\\
  \subfigure[Cosine]{\includegraphics[width=0.335\textwidth]{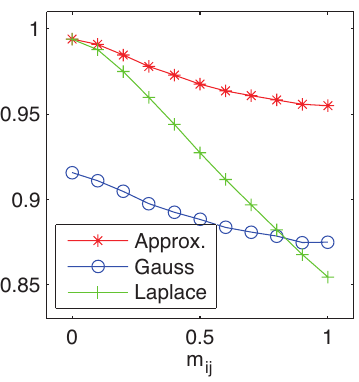}}
  \subfigure[KL
  divergence]{\includegraphics[width=0.32\textwidth]{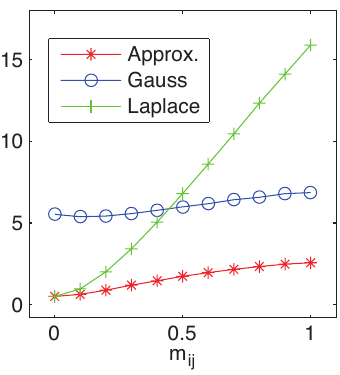}}
  \subfigure[Squared
  distance]{\includegraphics[width=0.325\textwidth]{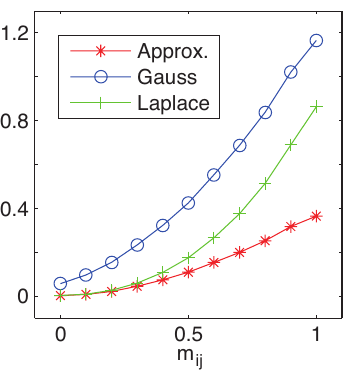}}
  \caption{\label{fig:comparison} Numerical validation of the
  approximation in (\ref{eqn:PDF}). (a) Contour plots of a simple
  non-parametric estimate of the density $p_s$. (b) Contour plot of the
  approximation $\tilde{p}_s$. (c-e) Goodness of the approximation as a
  function of $m_{12}$. ``Approx.'' is the approximation; ``Gauss'' and
  ``Laplace'' are standard models (multivariate Gaussian density, and
  Laplace factorizable density, respectively). Note that for (c), larger
  is better, while for (d) and (e), smaller is better.}
 \end{figure}
 Here, we investigate the validity of the approximative pdf
 $\tilde{p}_s(\vector{s};\m)$ in (\ref{eqn:PDF}) using numerical
 simulations.
 
 We generated a large number of samples for dependent non-Gaussian
 components according to the generative model in Section
 \ref{ssec:model}, fitted a nonparametric density to the sample, and
 compared it with our approximation in (\ref{eqn:PDF}). The dimension of
 $\vector{s}$ was $d=2$ and the size of the sample was $T=10^6$. The
 $u_{ij}$ were drawn from the inverse-Gamma distribution in
 (\ref{eqn:pij_def}) with $m_{11}=m_{22}=1$. We performed the comparison
 for multiple $m_{12}$ between 0 and 1. Using the generated sample, the
 density $p_s$ in (\ref{eqn:intgen}) was estimated as a normalized
 histogram. The approximative pdf $\tilde{p}_s(\vector{s},\m)$ was
 normalized using numerical integration and evaluated with the same
 $m_{ij}$ used to generate the sample.

 We evaluated the goodness of the approximation using three different
 measures,
 \begin{align}
  \text{ang}(p_s,\tilde{p}_s)&=
  \frac{\sum_{l=1}^N\sum_{k=1}^Np_s(l,k)\tilde{p}_s(l,k)}
  {\sqrt{\sum_{l=1}^N
  \sum_{k=1}^Np_s(l,k)^2\sum_{l=1}^N\sum_{k=1}^N\tilde{p}_s(l,k)^2}},
  \label{eqn:angle}\\ \text{KL}(p_s,\tilde{p}_s)&=
  \sum_{l=1}^N\sum_{k=1}^N\log\frac{p_s(l,k)}{\tilde{p}_s(l,k)}p_s(l,k),
  \label{eqn:KL}\\ \text{SQ}(p_s,\tilde{p}_s)&= \sum_{l=1}^N\sum_{k=1}^N
  \left\{p_s(l,k)-\tilde{p}_s(l,k)\right\}^2p_s(l,k), \label{eqn:SQ}
 \end{align}
 where $p_s(l,k)$ and $\tilde{p}_s(l,k)$ denote the values of the
 two-dimensional normalized histogram and the normalized
 $\tilde{p}_s(\vector{s};\m)$ in bin $(l,k)$, respectively. Equation
 (\ref{eqn:angle}) is the cosine of the angle between $p_s$ and
 $\tilde{p}_s$, and the larger the value, the better the
 approximation. Equations (\ref{eqn:KL}) and (\ref{eqn:SQ}) are the KL
 divergence and the expected squared distance, respectively. For
 comparison, we additionally computed these distance measures for a
 Laplace factorizable pdf with the same mean and marginal variance of
 the generated sources $\vector{s}$, and a Gaussian pdf with the same
 mean and covariance matrix.

 The logarithms of $p_s$ and $\tilde{p}_s$ for $m_{12}=0.95$ are shown
 in Figure~\ref{fig:comparison}(a) and (b), respectively. The two pdfs
 seem to have similar properties in terms of the heavy-tailed profiles
 and linear correlations. Figures~\ref{fig:comparison}(c,d,e) show that
 $\tilde{p}_s$ approximates $p_s$ better than the Laplace and Gaussian
 distributions for all $m_{12}$. This is due to the fact that our
 approximation captures both the heavy-tails of $p_s$ and its dependency
 structure. Thus, we conclude that our approximation $\tilde{p}_s$
 captures, at least qualitatively, the basic characteristics of the true
 distribution of the dependent non-Gaussian components.
 \section{Estimation of the Model} \label{sec:algo}
 In this section, we first show how to estimate the linear mixing model
 (\ref{eqn:datamodel}) and the dependency matrix $\M$ formed by $\m$
 using score matching, and then discuss important implementation
 details.
  \subsection{Using Score Matching for the Estimation}
  Approximating $p_s$ in~\eqref{eqn:px} with $\tilde{p}_s$ from
  (\ref{eqn:PDF}) yields an approximative pdf for the data as
  \begin{equation}
   p_x(\vector{x})\approx\tilde{p}_x(\vector{x};\W,\m)\propto\prod_{i=1}^d
  \exp\left(-m_{ii}|\vector{w}_i^{\top} \vector{x}| -\sum_{j>i}m_{ij}
  |\vector{w}_i^{\top}\vector{x}-\vector{w}_j^{\top}\vector{x}|\right)
  |\W|, \label{eqn:model}
  \end{equation}
  where $\vector{w}_i$ denotes the $i$-th row of $\W$. Both $\W$ and
  $\m$ are unknown parameters which we wish to estimate from a set of
  $T$ observations $\{\vector{x}_1, \ldots, \vector{x}_T\}$ of
  $\vector{x}$. A conventional approach for the estimation would consist
  in maximizing the likelihood. However, maximum likelihood estimation
  cannot be done here because $\tilde{p}_x(\vector{x};\W,\m)$ is an
  unnormalized model, only defined up to a proportionality factor
  (partition function) which depends on $\m$. To cope with such
  estimation problems, several methods have been
  proposed~\citep{hinton2002training,hyvarinen2005estimation,Gutmann2012a}
  (see \citet{Gutmann2013b} for a review paper).  One of the methods is
  score matching \citep{hyvarinen2005estimation} whose objective
  function for models from the exponential family is a quadratic form
  \citep[Section~4]{hyvarinen2007some}. If $\W$ is fixed, $\tilde{p}_x$
  in (\ref{eqn:model}) belongs to the exponential family, so that
  estimation of $\m$ given an estimate of $\W$ would be straightforward
  with score matching. This motivated us to estimate the model in
  (\ref{eqn:model}) by score matching, and to optimize its objective
  function by alternating between $\W$ and $\m$.
  
  We next derive the score matching objective function $J(\W,\m)$ for
  the joint estimation of $\W$ and $\m$, and then show how it is
  simplified when $\W$ is considered fixed. By definition of score
  matching \citep{hyvarinen2005estimation},
  \begin{align}
   J(\W,\m)&=\frac{1}{T}\sum_{t=1}^T\sum_{k=1}^d
   \frac{1}{2}\psi_k(\vector{x}_t;\W,\m)^2+\partial_k
   \psi_k(\vector{x}_t;\W,\m), \label{eqn:J}
  \end{align}
  where $\psi_k$ and $\partial_k \psi_k$ are the first- and second-order
  derivatives of $\log \tilde{p}_x$ with respect to the $k$-th
  coordinate of $\vector{x}$,
 \begin{align}
  \psi_k(\vector{x};\W,\m) &=\frac{\partial
  \log\tilde{p}_x(\vector{x};\W,\m)}{\partial x_k}, & \partial_k
  \psi_k(\vector{x};\W,\m) &=\frac{\partial^2
  \log\tilde{p}(\vector{x};\W,\m)}{\partial x_k^2}.
 \end{align}
 For our model in (\ref{eqn:model}), we have
 \begin{align}
  \psi_k(\vector{x};\W,\m) &=-\sum_{i=1}^d
  m_{ii}G^{\prime}(\vector{w}_i^{\top}\vector{x})w_{ik}
  -\sum_{j>i}m_{ij}
  G^{\prime}(\vector{w}_i^{\top}\vector{x}-\vector{w}_j^{\top}\vector{x})
  (w_{ik}-w_{jk}),\label{eqn:psik}\\ \partial_k \psi_k(\vector{x};\W,\m)
  &=-\sum_{i=1}^d
  m_{ii}G^{\prime\prime}(\vector{w}_i^{\top}\vector{x})w_{ik}^2
  -\sum_{j>i}m_{ij}
  G^{\prime\prime}(\vector{w}_i^{\top}\vector{x}-\vector{w}_j^{\top}\vector{x})
  (w_{ik}-w_{jk})^2,\label{eqn:partial_psik}
 \end{align}
 where the absolute value in (\ref{eqn:PDF}) is approximated by
 $|u|\approx G(u)=\log\cosh(u)$ for numerical stability,
 $G^{\prime}(u)=dG(u)/du=\tanh(u)$ and $G^{\prime\prime}(u)=\sech^2(u)$.
 
 Inspection of (\ref{eqn:psik}) and (\ref{eqn:partial_psik}) shows that
 $\psi_k$ and $\partial \psi_k$ are linear functions of the $m_{ii}$ and
 $m_{ij}$. If we consider $\W$ fixed, and let $\g^w_k(\vector{x})$ be
 the column vector formed by the terms in (\ref{eqn:psik}) which are
 multiplied by the elements of $\m$, and let $\h^w_k(\vector{x})$ be the
 analogous vector for (\ref{eqn:partial_psik}), we have
 $\psi_k(\vector{x};\W,\m) = \m^{\top} \g^w_k(\vector{x})$ and
 $\partial_k \psi_k(\vector{x};\W,\m) = \m^{\top}
 \h^w_k(\vector{x})$. The superscript ``$w$'' is used as a reminder that
 the vectors $\g_k^w(\vector{x})$ and $\h_k^w(\vector{x})$ depend on
 $\W$. For fixed $\W$, we can thus write $J(\W,\m)$ in (\ref{eqn:J}) as
 a quadratic form $J(\m|\W)$,
 \begin{equation}
  J(\m|\W)
   =\frac{1}{2}\vector{m}^{\top}\left(\frac{1}{T}\sum_{t=1}^T\sum_{k=1}^d 
				 \g_k^w(\vector{x}_t)\g_k^w(\vector{x}_t)^{\top}\right)\vector{m}+
 \vector{m}^{\top}\left(\frac{1}{T}\sum_{t=1}^T \sum_{k=1}^d
 \vector{h}^w_k(\vector{x}_t)\right). \label{eqn:Jm}
 \end{equation}
 For fixed $\W$, an estimate of $\m$ is obtained by minimizing
 $J(\m|\W)$ under the constraint that the $m_{ii}$ are positive and the
 other $m_{ij}$ are nonnegative.
 
 However, for estimation of $\W$, fixing $\m$ does not lead to an
 objective which takes a simpler form than $J(\W,\m)$ in (\ref{eqn:J}).
 Therefore, we optimize $J(\W,\m)$ by a simple gradient descent whose
 details are given below.

\subsection{Implementation Details}
We estimate the parameters $\W$ and $\m$ of our statistical model
(\ref{eqn:model}) for dependent non-Gaussian components by alternately
minimizing $J(\W,\m)$ in (\ref{eqn:J}) with respect to $\W$ and $\m$. We
next discuss some important details in this optimization scheme.

In our discussion of $\tilde{p}_s$ in (\ref{eqn:PDF}), we noted that
larger values of $m_{ij}$, $j>i$, indicate stronger (conditional)
correlation between components $s_i$ and $s_j$. In preliminary
simulations, we observed that, sometimes, the estimated $m_{ij}$ would
take much larger values than the estimated $m_{ii}$, leading to
estimated sources $\hat{s}_i$ and $\hat{s}_j$, and hence estimated
features $\widehat{\vector{w}}_i$ and $\widehat{\vector{w}}_j$, which were
almost the same. In order to avoid this kind of degeneracy, we imposed
the additional constraint that a $m_{ii}$ had to be larger than the
off-diagonal $m_{ij}$ summed together. Having this additional constraint,
we found that the strict positivity constraint on the $m_{ii}$ could
be relaxed to non-negativity. In summary, we imposed the following
constraints on $\m$:
\begin{align}
 (\forall (i,j): i\le j) \quad 0&\le m_{ij}, & (\forall i) \quad
 \sum_{j\neq i}m_{ij}&\leq m_{ii}. \label{eqn:constraints}
\end{align}
The constraints are linear, so that constrained minimization of
$J(\m|\W)$ can be done by standard methods from quadratic programming.

The mixing model (\ref{eqn:datamodel}) has a scale indeterminancy
because dividing a feature by some number while multiplying the
corresponding source by the same amount does not change the value of
$\vector{x}$. While this scale indeterminancy is a well-known phenomenon
in ICA, the situation is here more complicated because we have terms of
the form $m_{ii} |\vector{w}_i^\top\vector{x}|$ and $m_{ij} |\vector{w}_i^\top
\vector{x}-\vector{w}_j^\top\vector{x}|$ in the model-pdf instead of the more
simple $|\vector{w}_i^\top \vector{x}|$ terms found in ICA. While in ICA, the
scale indeterminancy is not a problem for maximum likelihood estimation,
we have found that for our model, it was necessary to explicitly resolve
the indeterminancy by imposing a unit norm constraint on the
$\vector{w}_i$ (for whitened data).

For the optimization of $J(\W,\m)$, we have to choose some initial
values for $\W$ and $\m$. We initialized $\W$ using a
maximum-likelihood-based ICA algorithm, additionally imposing the norm
constraint on the $\vector{w}_i$. In more detail, we initialized $\W$ as
$\widehat{\W}$,
\begin{align}
 \widehat{\W}&= \argmin_{1 \le i\le d, \|\vector{w}_i\|=1} J_0(\W),& J_0(\W) &=\frac{1}{T}\sum_{t=1}^T\sum_{i=1}^d
 G(\vector{w}_i^{\top}\vector{x}_t) -\log|\det\W|.\label{eqn:W0}
 \end{align}
Given the initial value $\widehat{\W}$, we obtained an initial value for
$\m$ by minimizing $J(\m|\widehat{\W})$ in (\ref{eqn:Jm}) under the
constraints in (\ref{eqn:constraints}).

While $J(\m|\W)$ can be minimized by quadratic programming, minimization
of $J_0(\W)$ and $J(\W,\m)$ for fixed $\m$ has to be done by less
powerful methods. We used a simple gradient descent algorithm where the
step-size $\mu_k$ at each iteration $k$ was chosen adaptively by trying
out $2 \mu_{k-1}$ and $1/2 \mu_{k-1}$ in addition to $\mu_{k-1}$, and
selecting the one which yielded the smallest objective.

Algorithm~1 summarizes our approach to estimate the model
(\ref{eqn:model}), where it is assumed that the data have already been
preprocessed by whitening and, optionally, dimension reduction by
principal component analysis (PCA). In the proposed method, good
initialization is important because the objective function has local
optima, which can produce spurious correlations in the estimated
$\vector{s}$. Therefore, we first perform ICA to give reasonable
initialization both for $\M$ and $\W$. One weakness of the proposed
method is that optimization for high-dimensional and large data can be
slow compared with ICA because we alternately repeat Step~1 and~2. To
alleviate this problem, in Section~\ref{sec:real}, we perform
dimensionality reduction by PCA, and estimate $\W$ and $\M$ based on a
randomly chosen subset of data samples at every repeat of Step~1 and~2.
\begin{myalgo}
 {\bf Algorithm~1: Estimation of the mixing matrix $\A$ and dependency
   matrix $\M$ } \newline \newline {\sf {\bf Input:} Data
   $\{\vector{x}_1,\vector{x}_2,\dots,\vector{x}_T\}$ which have been
   whitened.
 \begin{itemize}
  \item Initialization (ICA): Compute $\widehat{\W}$ as in
	(\ref{eqn:W0}). Fixing $\W = \widehat{\W}$, compute $\mhat$ by
	minimizing $J(\m|\W)$ in (\ref{eqn:Jm}) under the constraints in
	(\ref{eqn:constraints}), using a standard solver for quadratic
	programs.
	
  \item Repeat Step 1 and Step 2 until some conventional convergence
	criterion is met:
	 \begin{enumerate}
         \item[Step 1] Fixing $\m=\mhat$, update $\widehat{\W}$ by
		      taking one gradient step to minimize $J(\W, \m)$
		      in (\ref{eqn:J}) under the unit norm constraint on
		      the rows $\vector{w}_i$ of $\W$.
         \item[Step 2] Fixing $\W = \widehat{\W}$, re-compute $\mhat$ by
		      minimizing $J(\m|\W)$ in (\ref{eqn:Jm}) under the
		      constraints in (\ref{eqn:constraints}), using a
		      standard solver for quadratic programs.
	 \end{enumerate}
 \end{itemize}
 {\bf Output:} Mixing matrix $\widehat{\A}=\widehat{\W}^{-1}$,
 dependency matrix $\widehat{\M}$ formed by $\mhat$.}
\end{myalgo}
\section{Simulations on Artificial Data}
\label{sec:ide}
In this section, using artificial data, we evaluate how well the
proposed method identifies the sources and their dependency
structure. The proposed method is compared to ICA and CTA.
\subsection{Methods}
For our evaluation, we used data generated according to the model in
Section~\ref{ssec:model}. We considered both data with independent
components and data with components which had statistical dependencies
within certain blocks. The interest of using independent components
(sources) in the evaluation is to check that the model does not impose
dependencies among the estimates when the underlying sources are truly
independent. For the independent sources, the $u_{ii}$ in
(\ref{eqn:lambda}) were sampled from an inverse-Gamma distribution with
shape parameter $k_{ii}=2$ and scale parameter
$(m^{\prime}_{ii})^2=1$. The other elements $u_{ij}, i \neq j$, were set
to zero. Thus, $\Lam$ was a diagonal matrix, with $\lambda_{ii}=u_{ii}$,
and the generated sources $s_i$ were statistically independent on each
other. For the block-dependent sources, $\lambda_{ii}=\sum_{k=1}^d
u_{ik}$ and $\lambda_{ij}=-u_{ij}$ as in (\ref{eqn:lambda}).  The
$u_{ii}$ were for all $i$ from an inverse-Gamma distribution with shape
parameter $k_{ii}=2$ and scale parameter $(m^{\prime}_{ii})^2=1$. The
variables $u_{12}$, $u_{13}$ and $u_{23}$ were sampled from an
inverse-Gamma distribution with shape parameter $k_{ii}=2$ and scale
parameter $(m^{\prime}_{ij})^2=1/3$, while the remaining $u_{ij}$ were
set to zero. With this setup, the $s_1$, $s_2$ and $s_3$ are
statistically dependent while the other sources are conditionally
independent. This dependency structure creates a block structure in the
linear and energy correlation matrices where any pairs of
$(s_1,s_2,s_3)$ show relatively stronger dependencies than the other
pairs (Figure~\ref{fig:blocksource} (c) and (d)). The energy correlation
matrix is the correlation matrix of the squared random variables whose
$(i,j)$-th element is given by
\begin{align}
 \frac{E\{s_i^2s_j^2\}-E\{s_i^2\}E\{s_j^2\}}
 {\sqrt{E\{(s_i^2-E\{s_i^2\})^2\}E\{(s_j^2-E\{s_j^2\})^2\}}},
\end{align}
where $E$ denotes the expectation and $E\{s_i\}=0$. After generating the
sources, each component was standardized so that it has the zero mean
and unit variance.

The observed data were generated from the mixing model
(\ref{eqn:datamodel}) where the elements in $\A$ were sampled from the
standard normal distribution. The data dimension was $d=10$ and the
total number of observations was $T=20,000$. The preprocessing was
whitening based on PCA. The performance matrix $\P=\W\A$ was used to
visualize and evaluate the results. If $\P$ is close to a permutation
matrix, the sources are well-identified.

To measure the goodness of the estimated dependency matrix, we used the
scale parameters of the inverse-Gamma distributions employed to generate
the sources in this simulation. For independent sources, first we
constructed a reference matrix $\M^{\prime}$ by setting the diagonals
and off-diagonals to $m_{ii}^{\prime}=1$ and zeros, respectively. To
enable a comparison to the reference matrix, we normalize $\widehat{\M}$
by diving $\widehat{m}_{ij}$ by
$\sqrt{\widehat{m}_{ii}\widehat{m}_{jj}}$ so that the diagonals are all
ones, and denote the normalized $\widehat{\M}$ by
$\widehat{\M}^{\prime}$. Finally, the goodness was measured by
 \begin{align}
  \text{Error}_{\text{M}^{\prime}}=\|\M^{\prime}-\widehat{\M}^{\prime}\|_{\text{Fro}},
  \label{eqn:errorM}
 \end{align}
 where $\|\cdot\|_{\text{Fro}}$ denotes the Frobenius norm. The reason
 of the error definition (\ref{eqn:errorM}) is that since the different
 shape parameter $k_{ii}$ from the ones in the inverse-Gamma
 distribution (\ref{eqn:pij_def}) was used for numerical stability, we
 could not know the exact $\M$ and therefore had to focus on the
 relative values of the elements in $\widehat{\M}$. For block sources,
 we constructed the reference matrix by setting the diagonals in
 $\M^{\prime}$ to $1$, and the off-diagonals to
 $m_{ij}^{\prime}/\sqrt{\sum_{j=1}^dm_{ij}^{\prime}\sum_{i=1}^dm_{ji}^{\prime}}$
 inside the block and to zeros outside the block.  As a result,
 $\M^{\prime}$ becomes a diagonally dominant matrix with a block
 structure.

For comparison, we performed ICA and CTA~\citep{sasaki2013correlated} on
the same data.\footnote{The MATLAB package for CTA is available at the
first author's web page.}
The ICA method was the same as the method used to initialize $\W$ in
Algorithm~1, with the unit norm constraint on the rows $\vector{w}_i$ of
$\W$. For all methods, to avoid local optima, we performed $10$ runs
with different initialization of $\W$ and chose the run with the best
value of each objective function. For ICA and CTA, after estimating
$\W$, we estimated their dependency matrices by minimizing
(\ref{eqn:Jm}) with the same constraints (\ref{eqn:constraints}).
\subsection{Results}
\label{ssec:artres} 
\setcounter{subfigure}{0}
 \begin{figure}[!p]
  \centering \subfigure[Performance
  Matrices]{\includegraphics[width=0.75\textwidth]{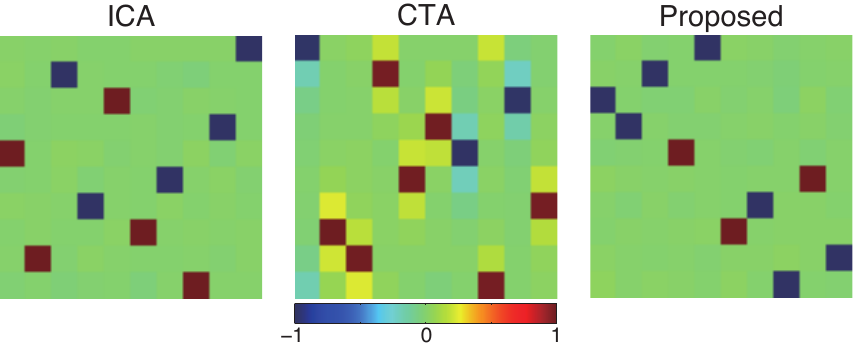}}
  \subfigure[Dependency
  Matrices]{\includegraphics[width=0.75\textwidth]{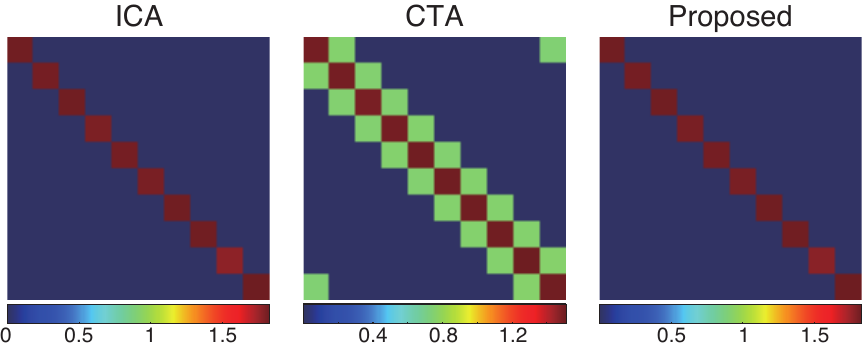}}
  \subfigure[Linear Correlation
  Matrices]{\includegraphics[width=0.95\textwidth]{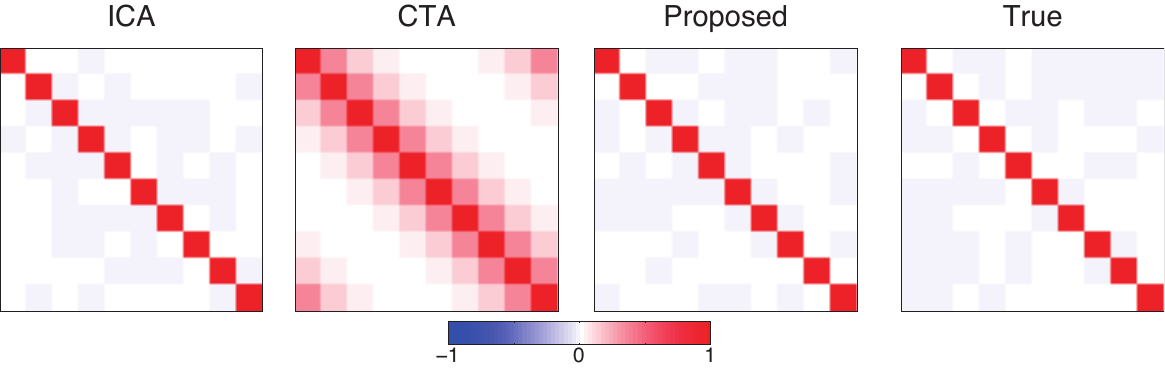}}
  \subfigure[Energy Correlation
  Matrices]{\includegraphics[width=0.95\textwidth]{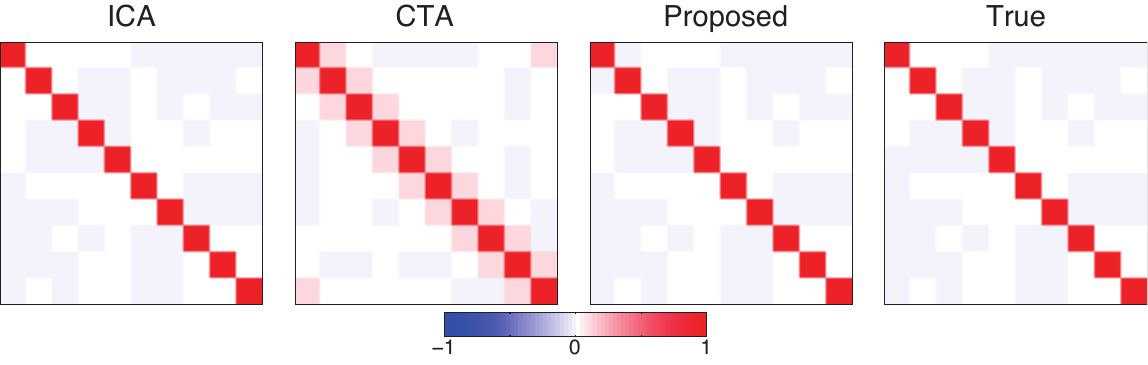}}
  \caption{\label{fig:indsource} Simulation results for independent
  sources. From (b) to (d), the permutation indeterminacy of ICA and the
  proposed method was compensated so that the largest element in each
  row of the performance matrix is on the diagonal. The true linear and
  energy correlation matrices of the sources are presented in the
  rightmost figures of (c) and (d).}
 \end{figure}
 \setcounter{subfigure}{0}
 \begin{figure}[!t]
  \vspace{5mm}
  \centering \subfigure[Performance
  Matrices]{\includegraphics[width=0.75\textwidth]{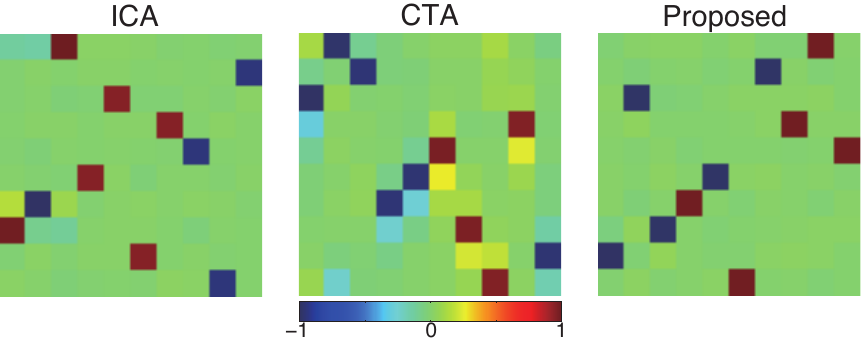}}
  \subfigure[Dependency
  Matrices]{\includegraphics[width=0.75\textwidth]{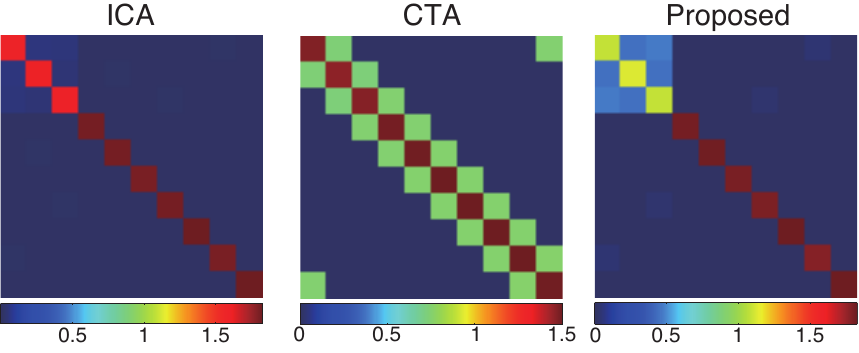}}
  \subfigure[Linear Correlation
  Matrices]{\includegraphics[width=0.95\textwidth]{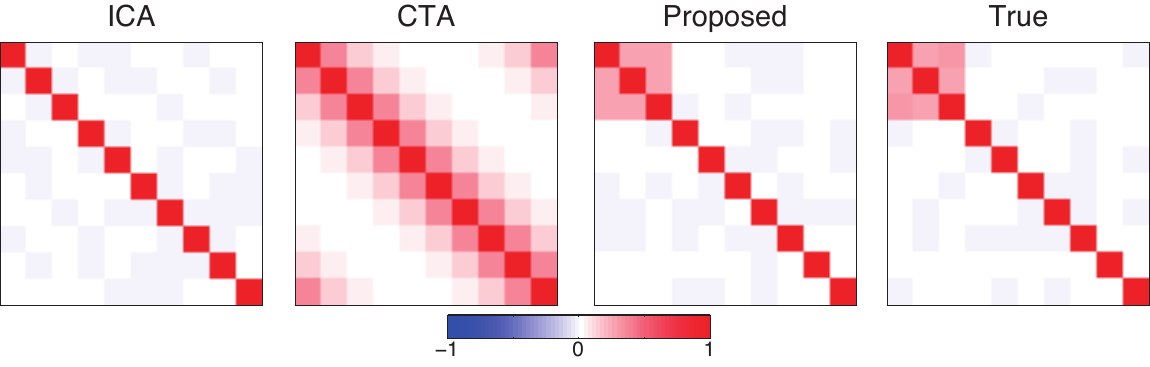}}
  \subfigure[Energy Correlation
  Matrices]{\includegraphics[width=0.95\textwidth]{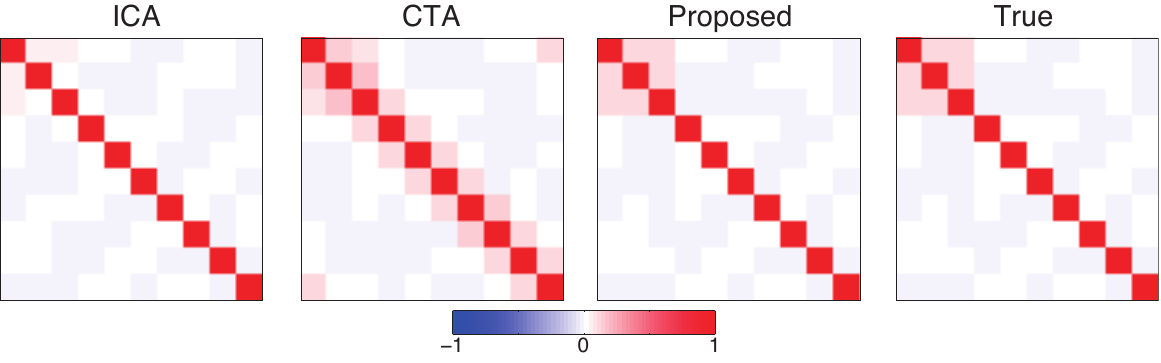}}
  \caption{\label{fig:blocksource} Simulation results for
  block-dependent sources. From (b) to (d), the permutation
  indeterminacy of ICA and the proposed method was compensated so that
  the largest element in each row of the performance matrix is on the
  diagonal.}
 \end{figure}
\setcounter{subfigure}{0}
 \begin{figure}[!t]
  \centering \subfigure[Amari
  Index]{\includegraphics[width=0.75\textwidth]{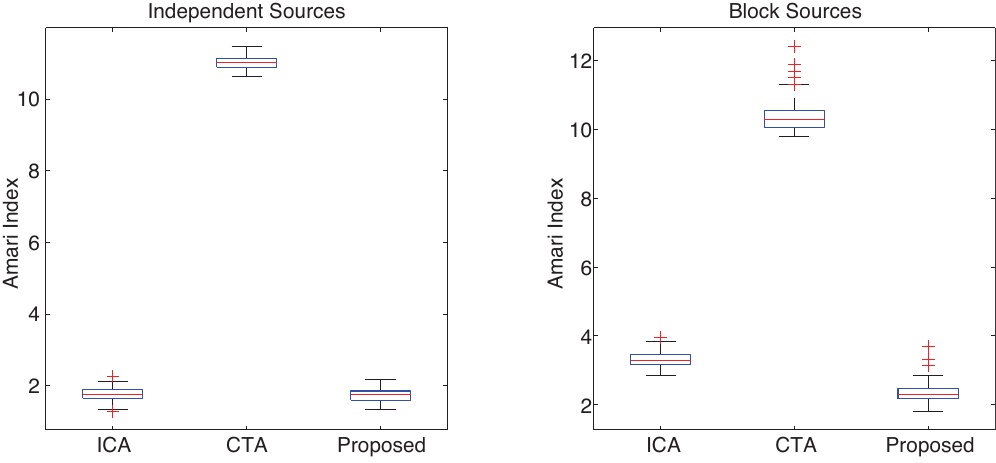}}
  \subfigure[$\text{Error}_{\text{M}^{\prime}}$
  ]{\includegraphics[width=0.75\textwidth]{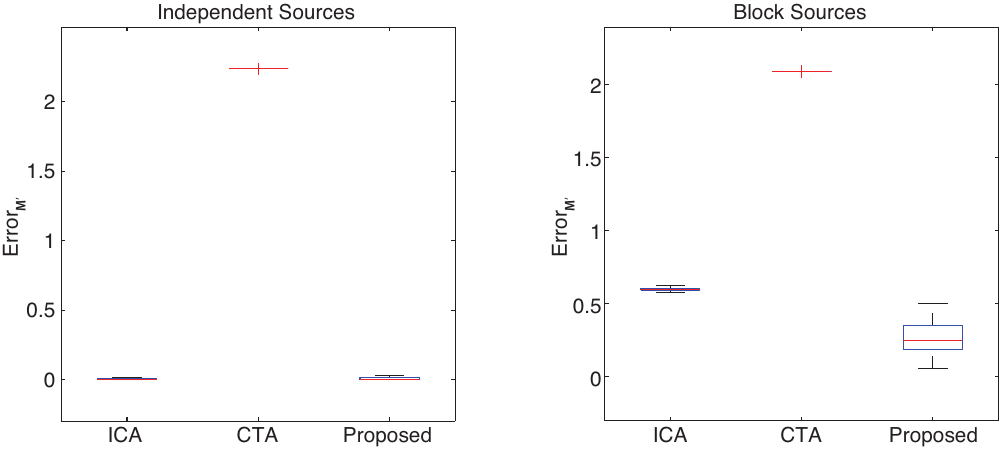}}\\
  \caption{\label{fig:Aind} Estimation error for $100$ runs, for both
  independent and block sources, summarized in terms of Amari index and
  Error$_{\text{M}^{\prime}}$. In the comparisons, ICA and CTA were used
  to estimate $\W$, and their dependency matrices were estimated by
  minimizing the objective in (\ref{eqn:Jm}) with the same constraints
  (\ref{eqn:constraints}).}
 \end{figure}
 
 We first report the performance on a single dataset, and then the
 average performance on $100$ randomly generated datasets. The results
 for the independent sources from the single dataset are presented in
 Figure~\ref{fig:indsource}. For ICA and the proposed method, the
 performance matrices are close to permutation matrices
 (Figure~\ref{fig:indsource}(a)), the estimated dependency matrices
 resemble a diagonal matrix (Figure~\ref{fig:indsource}(b)), as they
 should be, and the correlation matrices are almost diagonal
 (Figures~\ref{fig:indsource}(c) and (d)). For CTA, on the other hand,
 the performance matrix includes more cross-talk, the dependency matrix
 is tri-diagonal, and the linear and energy correlation matrices are
 clearly different from a diagonal matrix. These unsatisfactory results
 for CTA come from the fact that the dependency structure of CTA is
 pre-fixed, and thus CTA forcibly imposes linear correlations among the
 estimated neighboring components even though the original components
 are linearly uncorrelated. This drawback has been already reported
 by~\citet{sasaki2013correlated}. In contrast, the proposed method
 learned automatically that the sources are independent, and solved the
 identifiability issue of CTA.
 
 The results for block-dependent sources are shown in
 Figure~\ref{fig:blocksource}. The proposed method separates the
 sources, that is, estimates the linear components, with good accuracy,
 while ICA has more errors. The performance matrix for CTA includes
 again a lot of cross-talk (Figure~\ref{fig:blocksource}(a)). Regarding
 $\M$ and the correlations matrices, we compensated for the permutation
 indeterminacy of the sources for both ICA and the proposed method so
 that the largest element in each row of the performance matrix is on
 the diagonal. Figure~\ref{fig:blocksource}(b) shows that the proposed
 method yields a dependency matrix with a clearly visible block
 structure in the upper left corner, while ICA and CTA do not. In
 addition, the linear and energy correlation matrices have the block
 structure for the proposed method, while ICA and CTA do not produce it
 (Figure~\ref{fig:blocksource}(c) and (d)). Thus, only the proposed
 method was able to correctly identify the dependency structure of the
 latent sources.

 We further analyzed the average performance for $100$ randomly
 generated datasets. Figure~\ref{fig:Aind}(a) shows the distribution of
 the Amari index (AI)~\citep{amari1996} for the independent and
 block-dependent sources. AI is an established measure to assess the
 performance of blind source separation algorithms, and a smaller value
 means better performance. For independent sources, the proposed method
 has almost the same performance as ICA, while the performance of CTA is
 much worse. For block-dependent sources, the performance of the
 proposed method is better than ICA and CTA. This result means that when
 the original sources $s_i$ are linearly correlated, ICA is not the best
 method in terms of identifiability of the mixing matrix, and that
 taking into account the dependency structure for linear correlations
 improves the identifiability.

 For the goodness of the estimated dependency matrix, we plot the
 distribution of Error$_{\M^{\prime}}$ in Figure~\ref{fig:Aind}(b).  In
 this analysis, the permutation indeterminacy was compensated as done in
 Figure~\ref{fig:blocksource}. The plot confirms the qualitative
 findings in Figures~\ref{fig:indsource} and \ref{fig:blocksource}: the
 proposed method is able to capture the dependency structure of the
 sources better than ICA and CTA.
\section{Application to Real Data}
\label{sec:real}
This section demonstrates the applicability of the proposed method on
two kinds of real data: speech data and natural image data.

Basic ICA and related methods based on energy (square) correlation work
well on raw speech and image data, due to the symmetry of the data
distributions. The symmetry implies in particular that positive and
negative values of linear features are to some extent equivalent, as
implicitly assumed in computation of Fourier spectra or complex cell
outputs in models of early (mammalian) vision, which is compatible with
energy correlations.

However, on higher levels of feature extraction, such symmetry cannot be
found anymore, and energy correlations cannot be expected to be
meaningful. Our goal here is to apply our new method on such
higher-level features, where linear correlations are likely to be
important. In particular, we use speech spectrograms, and outputs of
complex cells simulating computations in the visual cortex,
respectively.
\subsection{Speech Data}
\label{ssec:SO}
\setcounter{subfigure}{0}
\begin{figure}
 \centering \subfigure[Estimated basis vectors (left) and dependency
 matrix (right) ]{\includegraphics[width=0.95\textwidth]{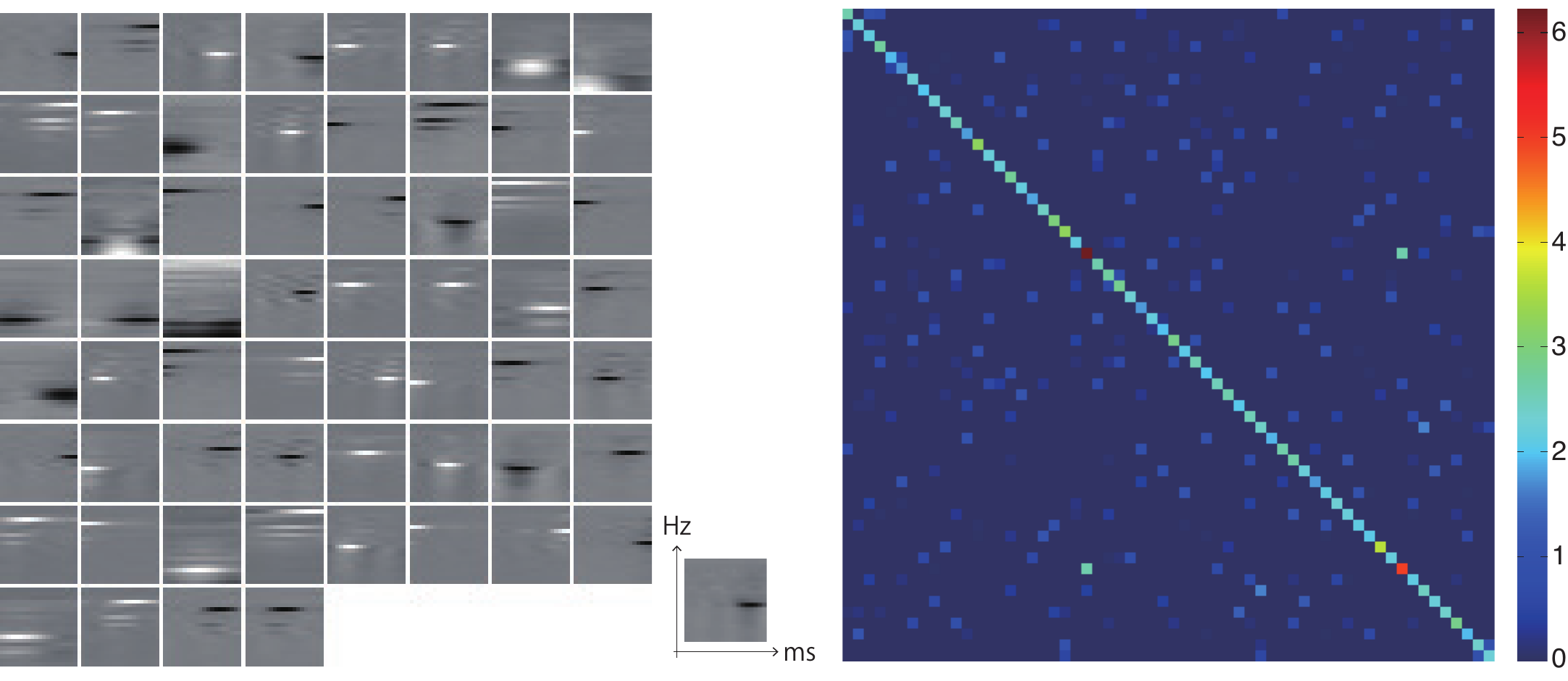}}
 \subfigure[For a selected $i$, the $m_{ij}$ as stem plots, and the
 corresponding basis vectors with strong $m_{ij}$ (the left-most being
 $\widehat{\vector{a}}_i$ and the others $\widehat{\vector{a}}_j$)
 ]{\includegraphics[width=0.95\textwidth]{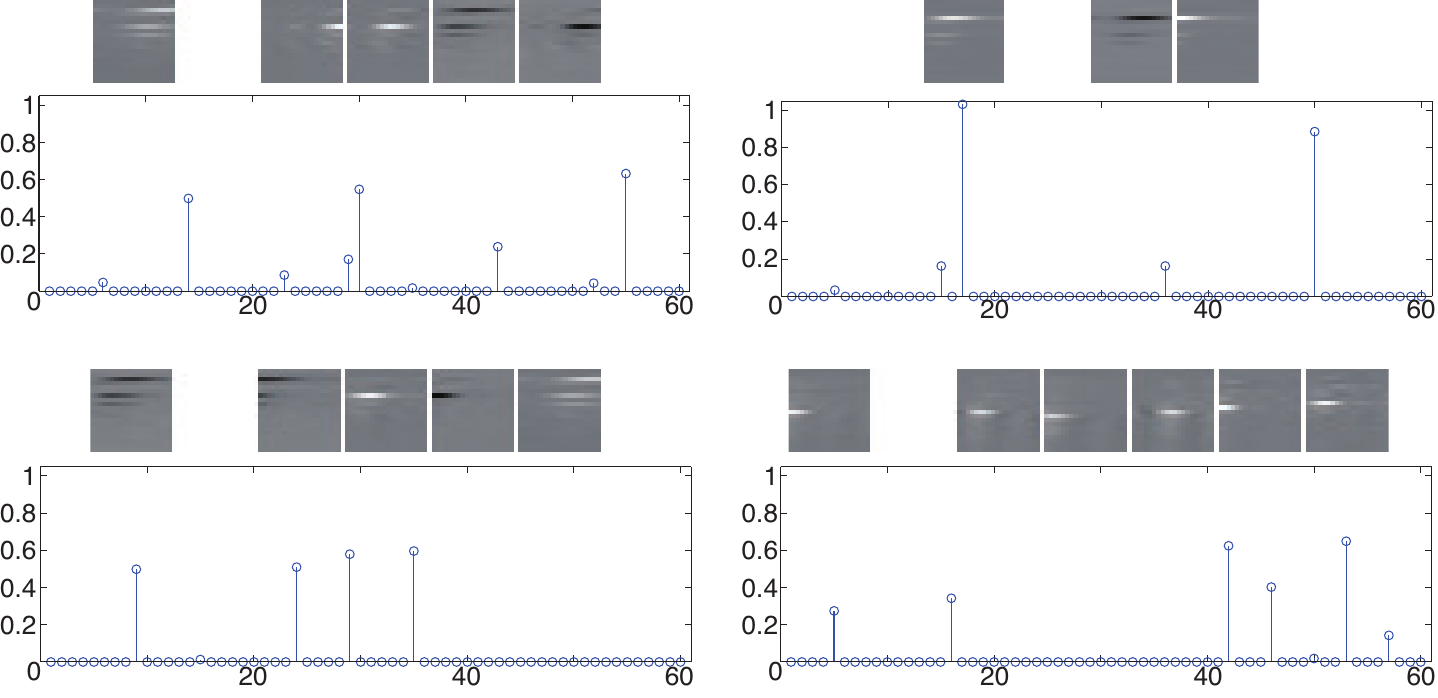}}
 \caption{\label{fig:SO} Results for speech (spectrogram) data.}
\end{figure}
\begin{figure}
 \centering \includegraphics[width=0.95\textwidth]{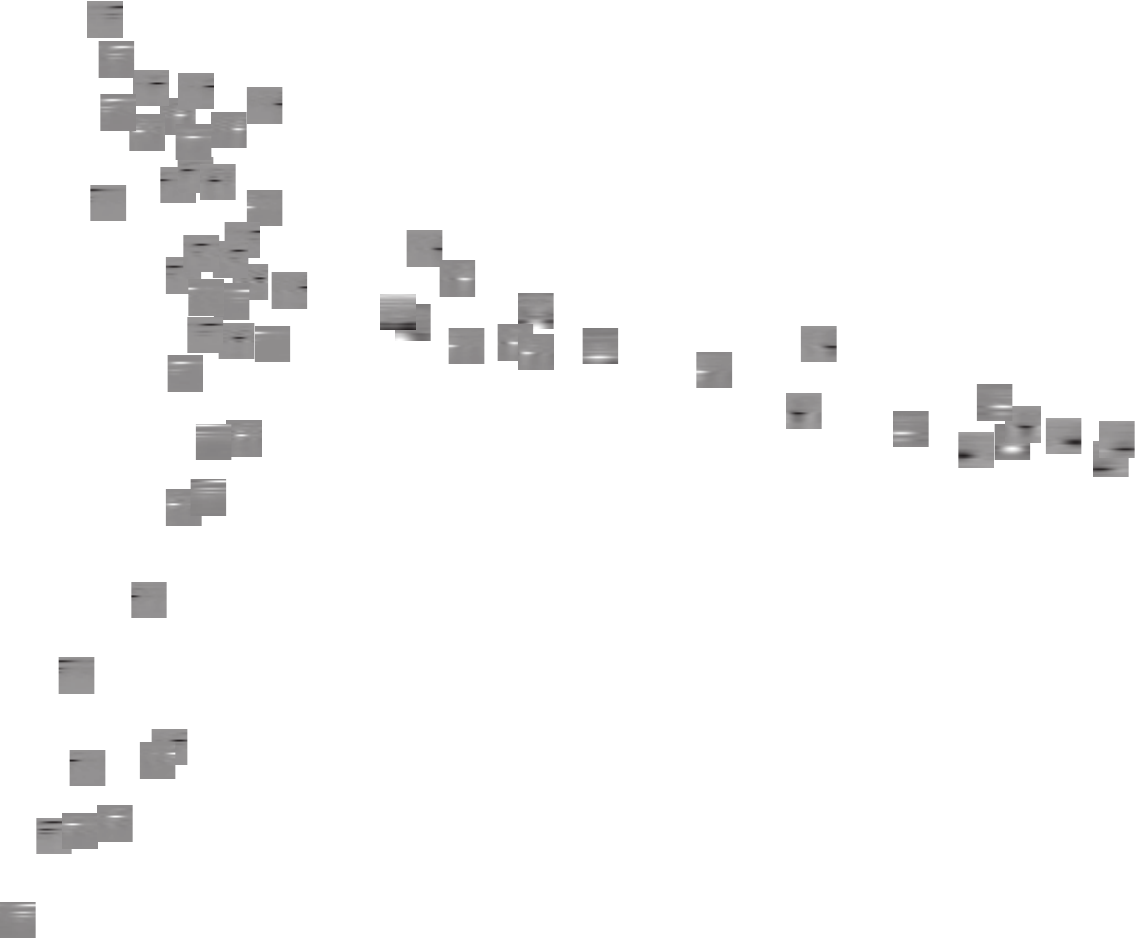}
 \caption{\label{fig:SO_graph} Speech data: Visualization of the
 estimated dependency structure between features (basis vectors) by
 MDS. In the figure, features with stronger $m_{ij}$ should be closer to
 each other in this visualization. The positions of some too close or
 too far-away features were magnified in order to show all features in a
 reasonable scale.}
\end{figure}
Previously, sparse coding~\citep{olshause1996emergence} and
ICA-related-methods have been applied to audio data to investigate the
basic properties of cells in the primary auditory cortex
(A1)~\citep{klein2003sparse,terashima2009sparse,terashima2013sparse}. More
recently, topographic ICA (TICA)~\citep{hyvarinen2001topographic} was
employed to analyze spectrogram data, and feature maps were learned
which are similar to the tonotopic maps in
A1~\citep{terashima2012topo}. However, in TICA, the dependency structure
is influenced by higher-order correlations only and it is fixed to
nearby components beforehand. Furthermore, using energy correlations for
spectrograms may not be well justified. Here, we lift these restrictions
and learn the dependency structure from the data by taking both linear
and higher-order correlations between the latent sources into account.

Following~\citet{terashima2012topo}, we used human narratives
data~\citep{international1999handbook}. The data were down-sampled to
$8$ kHz, and the spectrograms were computed by using the NSL
toolbox.\footnote{Available at
\url{http://www.isr.umd.edu/Labs/NSL/Downloads.html}} After re-sizing
the vertical (spectral) size of the spectrograms from $128$ to $20$,
short spectrograms were randomly extracted with the horizontal
(temporal) size equal to $20$. The vectorized spectrogram patches were
our $T=100,000$ input data points
$\{\vector{x}_1,\vector{x}_2,\dots,\vector{x}_T\}$.

As preprocessing, we removed the DC component of each $\vector{x}_t$,
and then rescaled each $\vector{x}_t$ to unit norm. Finally, whitening
and dimensionality reduction were performed simultaneously by PCA. We
retained $d = 60$ dimensions. To reduce the computational cost, in this
experiment, at every repeat of Step~1 and~2 in Algorithm~1, we randomly
selected $30,000$ data points from $T=100,000$ data points to be used
for estimation.

The estimated basis vectors $\vector{a}_i$ will be visualized in the
original domain as spectrograms. For the estimated dependency matrix, we
apply a multidimensional scaling (MDS) method to $\widehat{\M}$ to
visualize the dependency structure on the two-dimensional plane. To
employ MDS, we constructed a distance matrix from $\widehat{\M}$
similarly as done in~\citet{hurri2003temporal}: We first normalized each
element $\widehat{\m}_{ij}$ by
$\sqrt{\widehat{\m}_{ii}\widehat{\m}_{jj}}$ to make the diagonals ones,
then computed the square root of each element in the normalized matrix,
and finally subtracted each element from one. The purpose of MDS is to
project the points in a high-dimensional space to the two-dimensional
plane so that the distance in the high-dimensional space is preserved as
much as possible in the two-dimensional space. Thus, applying MDS should
yield a representation where the dependent features (basis vectors) are
close to each other.

The estimated basis vectors $\hat{\vector{a}}_i$ and dependency matrix
$\widehat{\M}$ are presented in Figure~\ref{fig:SO}(a). Most of the
basis vectors show vertically (spectrally) and horizontally (temporally)
localized patterns with single or multiple peaks. These properties have
been also found in previous work~\citep{terashima2012topo}. But, unlike
previous work, we also estimated the dependency structure from the
data. As shown in the right panel of Figure~\ref{fig:SO}(a), the
off-diagonal elements of the dependency matrix are sparse: most of the
elements are zero. Figure~\ref{fig:SO}(b) shows that basis vectors with
similar peak frequencies tend to have strong (conditional) dependencies.
The visualization of $\widehat{\M}$ further globally supports this
observation (Figure~\ref{fig:SO_graph}).

Compared with previous work, the properties of nearby features in
Figure~\ref{fig:SO_graph} seem to be more consistent: Nearby features
tend to have similar peak positions along with the spectral (vertical)
axis, while the peak positions on the temporal (horizontal) axis are
more random. On the other hand, \citet{terashima2012topo} found that
nearby features often show different peak positions on the spectral
axis, and the estimated features on the topographic map are locally
disordered. These results may reflect that linear correlations in sound
spectrogram data should be important dependencies, and that the proposed
method captured the structure in the data better than TICA.
\subsection{Outputs of Complex Cells}
\label{ssec:CC}
We next apply our method to the outputs of simulated complex cells in
the primary visual cortex when stimulated with natural image
data. Previously, ICA, non-negative sparse coding and CTA have been
applied to this kind of data, and some prominent features such as long
contours and topographic maps have been
learned~\citep{hoyer2002multi,hyvarinen2005statistical,sasaki2013correlated}.
However, these methods have either assumed that the features are
independent, or pre-fixed their dependency structure. Our method removes
this restriction and learns the dependency structure from the data.

As in the previous work cited above, we computed the outputs of the
simulated complex cells $\vector{x}$ as
\begin{align*}
  x'_k &= \left( \sum_{x,y}W_k^o(x,y)I(x,y) \right)^2+\left(
  \sum_{x,y}W_k^e(x,y)I(x,y) \right)^2, \nonumber\\ 
  x_k &= \log(x'_k+1.0), 
\end{align*}
where $I(x,y)$ is a $24\times 24$ natural image patch,\footnote{To
compute the complex cell outputs, we used the {\it contournet} package
which is available at
\url{http://www.cs.helsinki.fi/u/phoyer/software.html}.} and
$W_k^o(x,y)$ and $W_k^e(x,y)$ are odd- and even-symmetric Gabor
functions with the same spatial positions, orientation and
frequency. The total number of outputs was $T=100,000$. The complex
cells were arranged on a $6$ by $6$ spatial grid, having $4$
orientations each. In total, there were $144$ cells. Since the simulated
complex cells in this experiment are stimulated by natural images, we
regard this data as real data. We performed the same preprocessing steps
as in Section~\ref{ssec:SO} above; the dimensionality was here reduced
to $d=60$. As in the last section, to reduce the computational cost, we
randomly selected a subset of data points from the whole data points at
every repeat of the two steps in Algorithm~1. We visualized the basis
vectors as in previous
work~\citep{hoyer2002multi,hyvarinen2005statistical}: Each basis vector
is visualized by ellipses which have the orientation and spatial
position preferences of the underlying complex cells.

\begin{figure}[!t]
 \centering \subfigure[Estimated basis vectors (left) and dependency
 matrix (right)]{\includegraphics[width=0.95\textwidth]{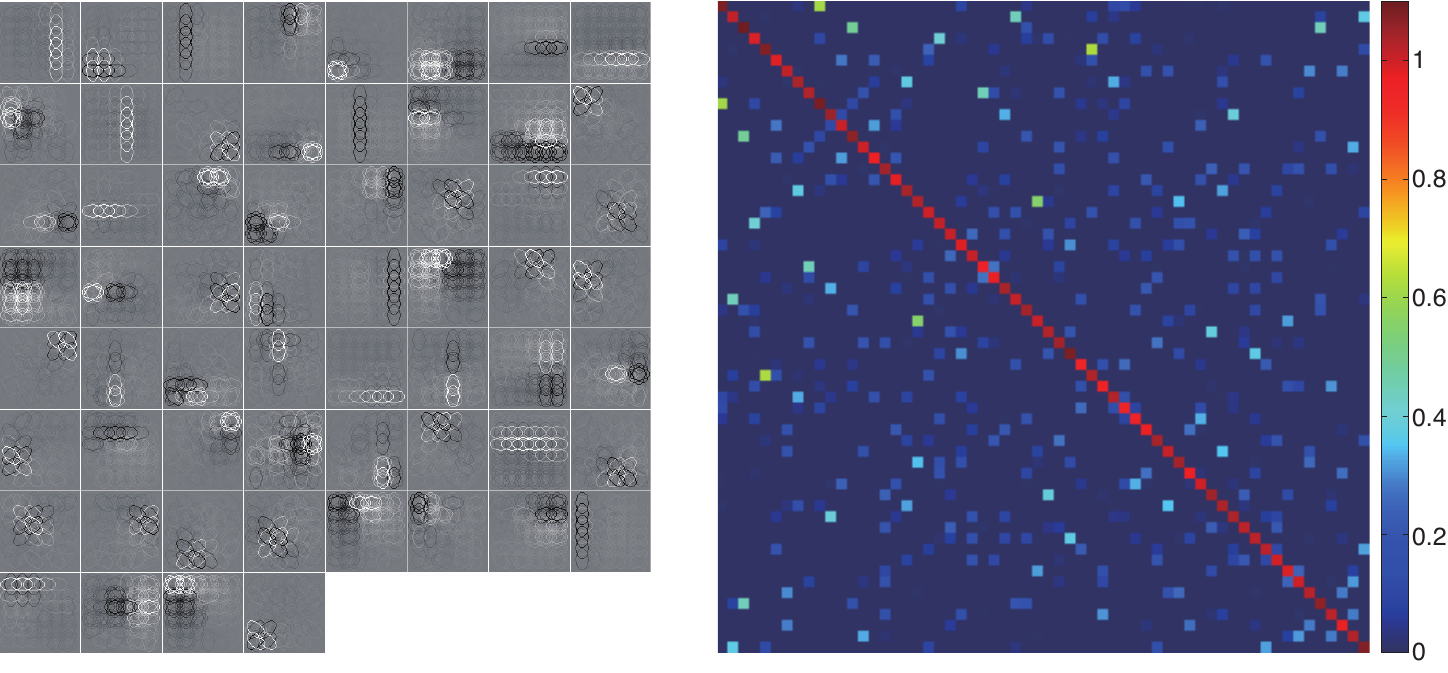}}
 \subfigure[For a selected $i$, the $m_{ij}$ as stem plots, and the
 corresponding basis vectors with stronger $m_{ij}$ (the left-most being
 $\widehat{\vector{a}}_i$ and the others
 $\widehat{\vector{a}}_j$)]{\includegraphics[width=0.95\textwidth]{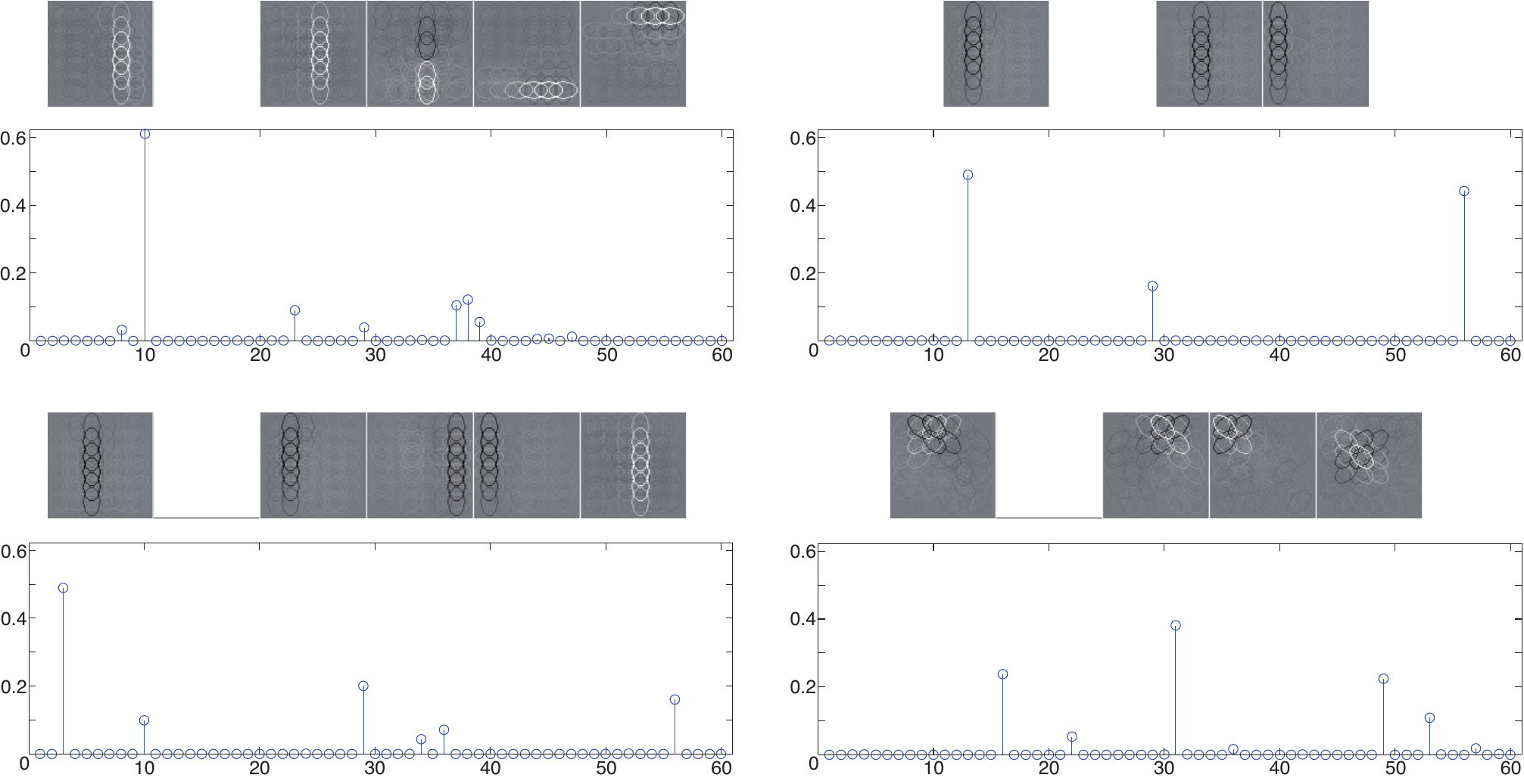}}
 \caption{\label{fig:CC} Results for the simulated complex cell data.}
\end{figure}

The estimated basis vectors $\hat{\vector{a}}_i$ and the dependency
matrix $\widehat{\M}$ are presented in Figure~\ref{fig:CC}. One
prominent kind of features among the basis vectors are long contours, as
also found in previous
work~\citep{hoyer2002multi,hyvarinen2005statistical}.

Unlike in previous work, we also learned the dependencies between the
features. As with the speech data, the off-diagonal elements of the
dependency matrix are sparse (Figure~\ref{fig:CC}(a), right), and
similar features tend to have stronger dependencies
(Figure~\ref{fig:CC}(b)).

\begin{figure}
 \centering \includegraphics[width=0.95\textwidth]{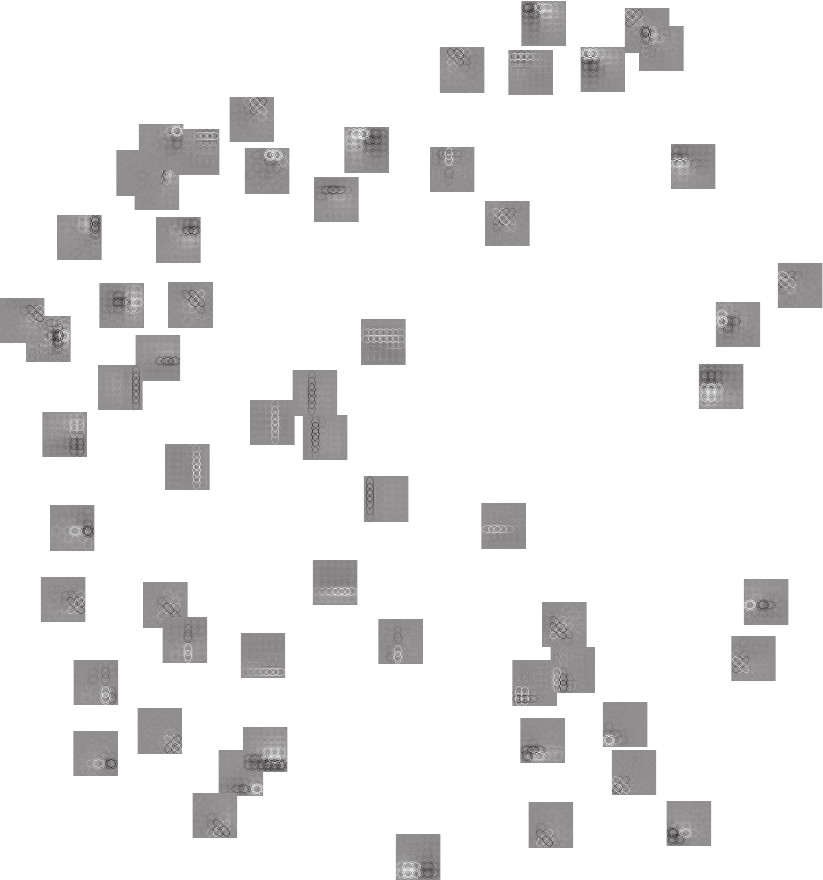}
 \caption{\label{fig:CC_graph} Complex cell outputs: Visualization of
 the estimated dependency structure between features (basis vectors) by
 MDS. In the graph, features with stronger $m_{ij}$ should be closer to
 each other in this visualization.}
\end{figure}

Figure \ref{fig:CC_graph} visualizes the dependency structure by MDS as
in Section~\ref{ssec:SO}. This visualization supports the observation
that the contour features tend to have stronger dependencies. In
particular, it is often the case that contour-features are closer to
other contour-features which are slightly shifted along their
non-preferred orientation. If put together, such two contour-features
would form either a broader contour of the same orientation or a
slightly bent even longer contour. This property is in line with
higher-level features learned using a three-layer model of natural
images~\citep{Gutmann2013}.

\begin{figure}[t]
 \centering \includegraphics[width=0.35\textwidth]{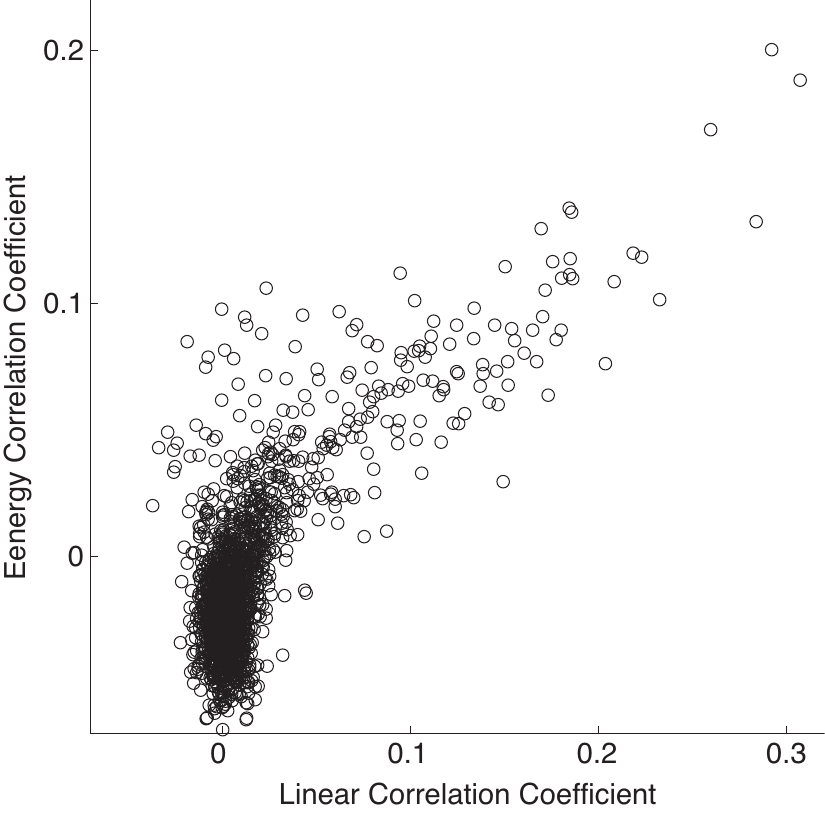}
 \caption{\label{fig:CC_scatter} A scatter plot for linear and energy
 correlation coefficients from the estimated sources. Each point
 corresponds to one pair of estimated components
 $(\widehat{s}_i,\widehat{s}_j)$.}
\end{figure}

We further investigate whether the proposed method estimated linearly
correlated components on real data in contrast to previous
energy-correlation-based
methods~\citep{karklin2005hierarchical,osindero2006topographic,koster2010two}.
Figure~\ref{fig:CC_scatter} shows a scatter plot for linear and energy
correlation coefficients for all the pairs of estimated components. As
the sparsity of $\widehat{\M}$ implies, most pairs have only weak
statistical dependencies. However, some pairs show both strong linear
and energy correlations. Thus, the proposed method did find linearly
correlated components on real data as well.

\setcounter{subfigure}{0}

\section{Discussion}
\label{sec:disc}
We discuss some connections to previous work and possible extensions of
the proposed method.
\subsection{Connection to Previous Work}
We proposed a novel method to simultaneously estimate non-Gaussian
components and their dependency (correlation) structure. So far, a
number of methods to estimate non-Gaussian components have been
proposed: ICA assumes that the latent non-Gaussian components are
statistically independent, and ISA~\citep{hyvarinen2000emergence} and
topographic
methods~\citep{hyvarinen2001topographic,mairal2011convex,sasaki2013correlated}
pre-fix the dependency structure inside pre-defined groups of components, or
overlapping neighborhoods of components. Methods to estimate
tree-dependency structures have also been proposed
\citep{bach2003beyond,Zoran09}. In contrast to these methods, our method
does not make any assumptions on the dependency structure to be
estimated.

Some methods based on two-layer models also estimate the dependency
structures of non-Gaussian components from
data~\citep{karklin2005hierarchical,osindero2006topographic,koster2010two}.
Most of the methods mainly focus on high-order correlations, assuming
that the components are linearly uncorrelated. The method proposed
by~\citet{osindero2006topographic} can estimate an overcomplete model as
well, which necessarily makes the estimated components linearly
correlated; however, it is unclear in what way such linear correlations
reflect the dependency structure of the underlying sources. Our method
models both linear and higher-order correlations explicitly. As a new
theoretical approach, in this paper, we proposed a generative model for
random precision matrices. The proposed generative model generalizes a
previous generative model for sources used in existing two-layer
methods~\citep{karklin2005hierarchical,osindero2006topographic}: The
previous model corresponds to a special case in our model where the
off-diagonal elements are deterministically zero.

Another line of related work is graphical models for latent
factors~\citep{NIPS2012_4636}. This work assumes that both the latent
factors and the undirected graph are sparse. The main goal of that
approach is to estimate a latent lower-dimensional representation of the
input data where pair-wise dependencies between latent factors are
represented by an undirected graph. The main difference to our work is
that \citet{NIPS2012_4636} use a constant precision matrix instead of a
stochastic one, and thus that they do not really model non-Gaussian
components. Instead, \citet{NIPS2012_4636} emphasize sparsity in the
sense that the undirected graphs to be estimated have sparse edges. As
discussed below, our method can be easily extended to include the
sparsity constraint while still keeping the objective function a simple
quadratic form.
\subsection{Extensions of the Proposed Method}
Next, we discuss some extensions of our model which could be
considered in future work.  One extension of the proposed method would
be to incorporate prior information or an additional constraint on
$\M$. Recently, a number of methods to estimate sparse Gaussian
graphical models has been
proposed~\citep{friedman2008sparse,banerjee2008model}. In our method,
the sparseness constraint on $m_{ij}$ can be easily incorporated into the
objective function $J(\m|\W)$ in (\ref{eqn:Jm}) yielding $J_\lambda(\m|\W)$,
\begin{align}
  J_\lambda(\m|\W)&= J(\m|\W) +\lambda \vector{m}^{\top}\vector{1},
\end{align}
where $\lambda$ is the regularization parameter, and $\vector{1}$ is a
vector of ones. The objective function $J_\lambda(\m|\W)$ is still a
quadratic form and its minimization is not more difficult than the
minimization of $J(\m|\W)$. However, this extension involves selecting
the parameter $\lambda$, which adds a complexity to the problem which
need to be solved. In practice, we may perform cross-validation (CV)
under some criterion, but employing CV is computationally demanding
especially in our alternate optimization. Therefore, selecting an
appropriate value of $\lambda$ is an issue in the future.

Another extension would be to estimate additional parameters modelling
linear correlations. As shown in Appendix~\ref{app:general}, the model
for the sources in (\ref{eqn:PDF}) can be generalized to
\begin{align*}
  \tilde{p}_s(\vector{s};\m,\The)&\propto \prod_{i=1}^d
  \exp\left(-\theta_{ii}m_{ii}|s_i| -\sum_{j>i}m_{ij}
  \sqrt{\theta_{ii}s_i^2+\theta_{jj}s_j^2+2\theta_{ij}s_is_j}\right),
\end{align*}
where the $\theta_{ii}$ are positive parameters and
$\theta_{ij}\in\left\{-1,0,1\right\}$ for $i\neq j$. The pdf in
(\ref{eqn:PDF}) is a special case of the above pdf: it is obtained for
$\theta_{ii}=1$ and $\theta_{ij}=-1$ for $i\neq j$.  Thus, by estimating
all the $\theta_{ij}$, we can estimate more complex dependency
structures, generalizing our method. However, estimating $\theta_{ij}$
leads a complicated optimization problem because the objective function
for score matching is no longer a quadratic form, and the $\theta_{ij}$
are discrete variables. That is why we decided to use the simpler model
in (\ref{eqn:PDF}) and leave this interesting challenge for future work.
\section{Conclusion}
\label{sec:conc}
In this paper, we proposed a method to simultaneously estimate
non-Gaussian components and their dependency structure. The dependency
structure is defined in terms of both linear and higher-order
correlations, can be represented in a convenient form, and thus can be
readily interpreted. Using score matching, the estimation of the
dependency structure is particularly simple because the objective
function takes a quadratic form and can be minimized using standard
methods from quadratic programming.

The proposed method generalizes previous methods: Independent component
and correlated topographic analysis both assume pre-fixed dependency
structures for the non-Gaussian components, while our method flexibly
estimates the dependency structure from the data themselves. Several
existing methods based on two-layer models also aim at estimating the
dependency structure, but they focus on higher-order correlations,
ignoring linear correlations.

We experimentally demonstrated on artificial data that the proposed
method improves identifiability of latent non-Gaussian components over
existing methods by learning their dependency structure, and
application to the outputs of simulated complex cells and the
spectrograms of natural audio data unveiled new kinds of relationships
among the latent non-Gaussian components.
 \subsection*{Acknowledgments}
 H. Sasaki carried out most of the work when he was a JSPS fellow at the
 University of Electro-Communications. The work was partly done when
 M. U. Gutmann was with the Department of Mathematics and Statistics,
 University of Helsinki, and supported by the Finnish Centre of
 Excellence in Computational Inference Research COIN. H. Shouno was
 partly supported by MEXT/KAKENHI JSPS KAKENHI Grant
 26120515. A. Hyv{\"a}rinen was supported by the Academy of Finland
 (Centre of Excellence in Inverse problems).

\appendix
\section*{Appendix}
\section{Calculations for (\ref{eqn:condPDF})}
\label{app:calc}
With (\ref{eqn:lambda}), the conditional pdf
(\ref{eqn:condPDFtmp}) can be written as
\begin{align}
  p(\vector{s}|\mathbf{U}) &=\frac{|\Lam|^{1/2}}{(2\pi)^{d/2}}
  \exp\left(-\frac{1}{2}\sum_{i=1}^d\left\{\lambda_{ii}s_i^2 +\sum_{j\neq
    i} \lambda_{ij}s_is_j\right\}\right), \\
  &=\frac{|\Lam|^{1/2}}{(2\pi)^{d/2}}
  \exp\left(-\frac{1}{2}\sum_{i=1}^d\left\{s_i^2\sum_{k=1}^du_{ik}
  -\sum_{j\neq i} s_is_ju_{ij}\right\}\right), \\
  &=\frac{|\Lam|^{1/2}}{(2\pi)^{d/2}}
  \exp\left(-\frac{1}{2}\sum_{i=1}^d\left\{ s_i^2u_{ii}+\sum_{k\neq
    i}s_i^2u_{ik} -\sum_{j\neq i}s_is_ju_{ij}\right\}\right).
\end{align}
Since the $u_{ij}$ are symmetric, we have further
\begin{align}
  p(\vector{s}|\mathbf{U}) &=\frac{|\Lam|^{1/2}}{(2\pi)^{d/2}}
  \exp\left(-\frac{1}{2}\sum_{i=1}^d\left\{ s_i^2u_{ii}+\sum_{k>i}
  \left(s_i^2+s_k^2\right)u_{ik} -2\sum_{j> i}
  s_is_ju_{ij}\right\}\right), \\
  &=\frac{|\Lam|^{1/2}}{(2\pi)^{d/2}}
  \exp\left(-\frac{1}{2}\sum_{i=1}^d\left\{s_i^2u_{ii} +\sum_{j>i}
  \left(s_i-s_j\right)^2u_{ij} \right\}\right),
\end{align}
which is (\ref{eqn:condPDF}). 
\section{Approximation of the Determinant in (\ref{eqn:approx})} 
\label{app:lower}
We prove that (\ref{eqn:approx}) approximates the determinant with a
lower bound and that the resulting approximation $\tilde{p}_s$ of
the source density $p_s$ is an unnormalized model.

By definition in (\ref{eqn:lambda}), $\Lam$ is a strictly diagonally
dominant matrix.  Ostrowski's inequality~\citep{ostrowski1937sur} thus
yields
\begin{align}
  |\Lam| &\geq \prod_{i=1}^d \left(\lambda_{ii}-\sum_{\substack{j=1\\j\neq
      i}}^d \lambda_{ij}\right),
\end{align}
and by definition of the $\lambda_{ij}$ in (\ref{eqn:lambda}),
we obtain
\begin{align}
  |\Lam| &\geq \prod_{i=1}^d  \left(\sum_{j=1}^du_{ij}-\sum_{\substack{j=1\\j\neq i}}^d
  u_{ij}\right)= \prod_{i=1}^d u_{ii},\label{eqn:lowerbound_approx}
\end{align}
which shows that the approximation (\ref{eqn:approx}) corresponds to a lower
bound of the determinant.

Applying the approximation of the determinant to $p(\vector{s}|\U)$ in (\ref{eqn:condPDF})
gives $\tilde{p}(\vector{s}|\U)$,
\begin{align}
  \tilde{p}(\vector{s}|\U)&=\frac{\prod_{i=1}^d \sqrt{u_{ii}}}{(2\pi)^{d/2}}
  \exp\left(-\frac{1}{2}\sum_{i=1}^d \left\{s_i^2u_{ii} +\sum_{j> i}
  \left(s_i-s_j\right)^2u_{ij} \right\}\right),
\end{align}
which is a lower bound for $p(\vector{s}|\U)$ due to (\ref{eqn:lowerbound_approx}),
\begin{equation}
  \tilde{p}(\vector{s}|\U) \le p(\vector{s}|\U).
\end{equation}
Using the approximation $\tilde{p}(\vector{s}|\U)$ instead of
$p(\vector{s}|\U)$ in (\ref{eqn:intgen}) yields an approximation
$\tilde{p}_s$ of the pdf of the dependent non-Gaussian components
which inherits the lower bound property,
\begin{equation}
  \tilde{p}_s(\vector{s}) = \int_{0}^{\infty}
  \tilde{p}(\vector{s}|\U)p_u(\U) d\U \le \int_{0}^{\infty}
  p(\vector{s}|\U)p_u(\U) d\U = p_s(\vector{s}).
\end{equation}
Since $p_s$ integrates to one, we have
\begin{equation}
  \int  \tilde{p}_s(\vector{s}) d \vector{s} \le 1,
\end{equation}
which means that $\tilde{p}_s$ is an unnormalized model. 
\section{Properties of the Nonlinearities in (\ref{eqn:g_def}) }
\label{app:logconvex}
We here investigate properties of the nonlinearities $g_{ij}(v)$ in
(\ref{eqn:g_def}). It is shown that, unless constant, the $\log
g_{ij}(v)$ are monotonically decreasing convex functions.

For the analysis, it is helpful to introduce the functions
$\phi_{ij}(u)$, $u\ge0$,
\begin{equation}
  \phi_{ij}(u) = \begin{cases} \sqrt{u} p_{ii}(u) & \mbox{if
    } i=j \\
    p_{ij}(u)& \mbox{otherwise},
  \end{cases}
\end{equation}
which allow us to rewrite the $g_{ij}(v)$ as
\begin{equation}
  g_{ij}(v) =  \int_0^{\infty} \exp\left (-\frac{1}{2} v u\right)\phi_{ij}(u) d u, \quad v \ge 0,
\end{equation}
for all $(i,j)$. The functions $\phi_{ij}$ are non-negative but they
are not pdfs if $i=j$ because they do not integrate to one. We note
that $g_{ij}$, $i\neq j$, is constant if $p_{ij}$ corresponds to a
Dirac measure concentrated at zero, that is, to a $u_{ij}$ which is
deterministically zero. In what follows, we assume that the $p_{ij}$
are pdfs with nonzero variance.

The derivative of $\log g_{ij}(v)$ is
\begin{eqnarray}
  \frac{\partial \log g_{ij}(v)}{\partial v} &=& \frac{ \frac{\partial
      g_{ij}(v)}{\partial v}} {g_{ij}(v)}\\
  &=& \frac{ -\frac{1}{2} \int_0^{\infty}  u\exp\left(-\frac{1}{2} v u\right)\phi_{ij}(u) d u}{ \int_0^{\infty} \exp\left(-\frac{1}{2} v u\right)\phi_{ij}(u) d u },
\end{eqnarray}
which is negative for all $v >0$. Importantly, the derivative is related
to the expected value $\mu_{ijv}$ of the pdf $\gamma_{ijv}(u)$,
\begin{equation}
  \gamma_{ijv}(u) =  \frac{\exp\left(-\frac{1}{2} v u\right)\phi_{ij}(u)}{\int_0^{\infty} \exp\left (-\frac{1}{2} v u\right)\phi_{ij}(u) d u },
\end{equation}
that is,
\begin{equation}
  \frac{\partial \log g_{ij}(v)}{\partial v} = -\frac{1}{2} \mu_{ijv}.
\end{equation}

The second derivative equals
\begin{eqnarray}
  \frac{\partial^2 \log g_{ij}(v)}{\partial v^2} &=& \frac{\partial^2 g_{ij}(v)}{\partial v^2}\frac{1}{g_{ij}(v)} - \left( \frac{\partial
    g_{ij}(v)}{\partial v}\right)^2 \frac{1}{g_{ij}(v)^2}\\
  &=& \frac{\partial^2 g_{ij}(v)}{\partial v^2}\frac{1}{g_{ij}(v)} -
  \left(\frac{\partial \log g_{ij}(v)}{\partial v}\right)^2\\
  &=& \frac{\partial^2 g_{ij}(v)}{\partial v^2}\frac{1}{g_{ij}(v)} -
  \frac{1}{4}\mu_{ijv}^2,
\end{eqnarray}
which can be written in terms of the variance of the pdf
$\gamma_{ijv}$: We have
\begin{equation}
  \frac{\partial^2 g_{ij}(v)}{\partial v^2} = \frac{1}{4} \int_0^{\infty}
  u^2\exp\left(-\frac{1}{2} v u\right)\phi_{ij}(u) d u
\end{equation}
so that
\begin{equation}
  \frac{\partial^2 g_{ij}(v)}{\partial v^2}\frac{1}{g_{ij}(v)} =
  \frac{1}{4}  \int_0^{\infty} u^2 \gamma_{ijv}(u) d u,
\end{equation}
and
\begin{equation}
  \frac{\partial^2 \log g_{ij}(v)}{\partial v^2} = \frac{1}{4}\left( \int_0^\infty u^2  \gamma_{ijv}(u) d u - \mu_{ijv}^2\right).
\end{equation}
The term in the parentheses is the variance of the pdf $\gamma_{ijv}$
and hence positive since the pdf is not degenerate. In conclusion, for
non-constant functions $g_{ij}$, we have
\begin{eqnarray}
  \frac{\partial \log g_{ij}(v)}{\partial v} &<&0,\\
  \frac{\partial^2 \log g_{ij}(v)}{\partial v^2} &>&0,
\end{eqnarray}
for all $v>0$, which makes the $\log g_{ij}$ monotonically decreasing convex functions.

\section{Inverse-Gamma Distributions and the Nonlinearities in (\ref{eqn:chosen_g}) }
\label{app:integral}
We here show that the functions $g_{ij}$ in (\ref{eqn:chosen_g}) are
obtained from inverse-Gamma distributed $u_{ij}$ with the pdfs in
(\ref{eqn:pij_def}). Inserting the pdf $p_{ii}$ from
(\ref{eqn:pij_def}) into (\ref{eqn:g_def}) yields $g_{ii}(s_i^2)$,
\begin{align}
  g_{ii}(s_i^2) &= \int_0^{\infty} \sqrt{u_{ii}}
  \exp\left(-\frac{1}{2} s_i^2 u_{ii} \right) p_{ii}(u_{ii}) d
  u_{ii}\\
  &= \int_0^{\infty} \sqrt{u_{ii}}
  \exp\left(-\frac{1}{2} s_i^2 u_{ii} \right)
  \frac{m_{ii}^2}{2}u_{ii}^{-2}
  \exp\left(-\frac{m^2_{ii}}{2u_{ii}}\right) d u_{ii}\\
  &= \frac{m_{ii}^2}{2} \int_0^{\infty} u_{ii}^{-3/2}\exp\left(-\frac{1}{2}\left[s_i^2 u_{ii} +\frac{m^2_{ii}}{u_{ii}}\right] \right)d u_{ii},
\end{align}
and the pdf $p_{ij}$ yields $g_{ij}((s_i-s_j)^2)$,
\begin{align}
 g_{ij}((s_i-s_j)^2) &= \int_0^{\infty} \exp\left(-\frac{1}{2}
 \left(s_i-s_j\right)^2u_{ij} \right) p_{ij}(u_{ij}) d u_{ij}\\ &=
 \int_0^{\infty} \exp\left(-\frac{1}{2} \left(s_i-s_j\right)^2u_{ij}
 \right) \frac{m_{ij}}{\sqrt{2\pi}} u_{ij}^{-3/2}
 \exp\left(-\frac{m^2_{ij}}{2u_{ij}}\right)d u_{ij}\\ &=
 \frac{m_{ij}}{\sqrt{2\pi}} \int_0^{\infty} u_{ij}^{-3/2}
 \exp\left(-\frac{1}{2}\left[(s_i-s_j)^2 u_{ij}
 +\frac{m^2_{ij}}{u_{ij}}\right] \right)d u_{ij}.
\end{align}
It can be seen that both $g_{ii}$ and $g_{ij}$ are defined in terms of
the integral
\begin{displaymath}
 \int_{0}^{\infty} x^{-3/2}
 \exp\left(-\frac{1}{2}\left[\beta^2x+\frac{\alpha^2}{x}\right]\right)dx,
\end{displaymath}
where $\alpha$ corresponds to $m_{ii}>0$ or $m_{ij}>0$, and $\beta$ to
$|s_i|$ or $|s_i-s_j|$. The integral can be solved in closed form
\citep[Equation~(47)]{sasaki2013correlated},
\begin{align}
  \int_{0}^{\infty}
  x^{-3/2}\exp\left(-\frac{1}{2}\left[\beta^2x+\frac{\alpha^2}{x}\right]\right)dx
  &= \left(\frac{2\pi}{\alpha^2}\right)^{1/2}\exp(-|\alpha\beta|),
  \label{eqn:formula2}
\end{align}
so that 
\begin{align}
  g_{ii}(s_i^2) &\propto \exp(-m_{ii} |s_i|)\\
  g_{ij}((s_i-s_j)^2) &\propto \exp(-m_{ij}|s_i-s_j|),
\end{align}
or, with $v>0$, 
\begin{align}
  \log g_{ii}(v) &= -m_{ii} \sqrt{v} +\mathrm{const},\\
  \log g_{ij}(v) &= -m_{ij} \sqrt{v} + \mathrm{const},
\end{align}
as claimed in the main text.
\section{A General Generative Model for Precision Matrices}
\label{app:general}
We here extend the generative model for precision matrices
(\ref{eqn:lambda}) to the more general form
\begin{align}
  \Lam&=\bfOme\circ\The, \label{eqn:genemodel2}
\end{align}
where $\circ$ denotes the Hadamard product or element-wise
multiplication. The matrix $\bfOme$ is symmetric and its elements are
non-negative random variables, $\The$ is a $d$ by $d$ deterministic symmetric matrix
whose diagonal elements $\theta_{ii}$ are positive and whose
off-diagonal elements $\theta_{ij}$, $i\neq j$, take values in
$\{-1,0,1\}$. The matrix $\The$ determines whether two components are
positively, not at all, or negatively conditionally correlated. The
matrix $\bfOme$ scales the variances and correlations randomly and is
defined as
\begin{align}
  \omega_{ij}=\left\{
  \begin{array}{cc}
    \sum_{k=1}^d u_{ik}, & i=j,\\
    u_{ij}, & i\neq j,
  \end{array}
  \right.\label{eqn:omega}
\end{align}
where the $u_{ij}=u_{ji}$ are symmetric non-negative random variables,
with $u_{ii}>0$. The model (\ref{eqn:genemodel2}) generalizes
the model (\ref{eqn:lambda}) of $\Lam$ in the main text which is
recovered for $\theta_{ii}=1$ and $\theta_{ij}=-1$.

From the discussion of (\ref{eqn:lambda}), it follows that $\bfOme$ is
always a strictly diagonally dominant matrix. We now show that $\Lam$
defined in (\ref{eqn:genemodel2}) has
the same property.
\begin{Prop}
 For $\bfOme$ as in (\ref{eqn:omega}), if $\theta_{ii}\geq|\theta_{ij}|$
 for all $i$ and $j$, $\Lam$ is a strictly diagonally dominant matrix.
\end{Prop}
\begin{proof}
 It is sufficient to prove that for all $i$, $\lambda_{ii}>\sum_{j\neq
 i}|\lambda_{ij}|$.  We compute
 \begin{align*}
  \lambda_{ii}-\sum_{\substack{j=1\\j\neq
  i}}^d|\lambda_{ij}|&=\theta_{ii}\omega_{ii}-\sum_{\substack{j=1\\j\neq
  i}}^d|\theta_{ij}|\omega_{ij}, \\
  &=\theta_{ii}\sum_{j=1}^du_{ij}-\sum_{\substack{j=1\\j\neq
        i}}^d|\theta_{ij}|u_{ij}, \\
    &=\theta_{ii}u_{ii}+\sum_{\substack{j=1\\j\neq
        i}}^d(\theta_{ii}-|\theta_{ij}|)u_{ij}.
 \end{align*}
  Since $\theta_{ii}>0$, $u_{ii} >0$, and $u_{ij}\ge 0$, the proposition
  follows.
\end{proof}
Since $\Lam$ is symmetric and strictly diagonally dominant, it is also
invertible and positive definite~\citep[Theorem 6.1.10]{horn1985matrix}.

Following the same procedure as in Section~\ref{ssec:derive}, we can
derive the following approximation of the pdf of the dependent
non-Gaussian components,
\begin{align}
  \tilde{p}_s(\vector{s};\m,\The)&\propto \prod_{i=1}^d
  \exp\left(-\theta_{ii}m_{ii}|s_i| -\sum_{j>i}m_{ij}
  \sqrt{\theta_{ii}s_i^2+\theta_{jj}s_j^2+2\theta_{ij}s_is_j}\right).
  \label{eqn:G-PDF}
\end{align}
For $\theta_{ii}=1$ and $\theta_{ij}=-1$, we recover the model in
(\ref{eqn:PDF}). However, estimation of the parameters is much more
difficult because (\ref{eqn:G-PDF}) no longer belongs to the
exponentially family and the $\theta_{ij}$ are discrete variables.

\bibliographystyle{apacite}
\bibliography{papers.bib}

\begin{thebibliography}{}

\bibitem [\protect \citeauthoryear {%
Amari%
, Cichocki%
\BCBL {}\ \BBA {} Yang%
}{%
Amari%
\ \protect \BOthers {.}}{%
{\protect \APACyear {1996}}%
}]{%
amari1996}
\APACinsertmetastar {%
amari1996}%
\begin{APACrefauthors}%
Amari, S.%
, Cichocki, A.%
\BCBL {}\ \BBA {} Yang, H.%
\end{APACrefauthors}%
\unskip\
\newblock
\APACrefYearMonthDay{1996}{}{}.
\newblock
{\BBOQ}\APACrefatitle {A new learning algorithm for blind signal separation} {A
  new learning algorithm for blind signal separation}.{\BBCQ}
\newblock
\BIn{} \APACrefbtitle {Advances in Neural Information Processing Systems}
  {Advances in neural information processing systems}\ (\BVOL~8, \BPGS\
  757--763).
\PrintBackRefs{\CurrentBib}

\bibitem [\protect \citeauthoryear {%
Bach%
\ \BBA {} Jordan%
}{%
Bach%
\ \BBA {} Jordan%
}{%
{\protect \APACyear {2003}}%
}]{%
bach2003beyond}
\APACinsertmetastar {%
bach2003beyond}%
\begin{APACrefauthors}%
Bach, F.%
\BCBT {}\ \BBA {} Jordan, M.%
\end{APACrefauthors}%
\unskip\
\newblock
\APACrefYearMonthDay{2003}{}{}.
\newblock
{\BBOQ}\APACrefatitle {Beyond independent components: trees and clusters}
  {Beyond independent components: trees and clusters}.{\BBCQ}
\newblock
\APACjournalVolNumPages{Journal of Machine Learning Research}{4}{}{1205--1233}.
\PrintBackRefs{\CurrentBib}

\bibitem [\protect \citeauthoryear {%
Ball{\'e}%
, Laparra%
\BCBL {}\ \BBA {} Simoncelli%
}{%
Ball{\'e}%
\ \protect \BOthers {.}}{%
{\protect \APACyear {2015}}%
}]{%
balle2015density}
\APACinsertmetastar {%
balle2015density}%
\begin{APACrefauthors}%
Ball{\'e}, J.%
, Laparra, V.%
\BCBL {}\ \BBA {} Simoncelli, E\BPBI P.%
\end{APACrefauthors}%
\unskip\
\newblock
\APACrefYearMonthDay{2015}{}{}.
\newblock
{\BBOQ}\APACrefatitle {Density modeling of images using a generalized
  normalization transformation} {Density modeling of images using a generalized
  normalization transformation}.{\BBCQ}
\newblock
\APACjournalVolNumPages{arXiv preprint arXiv:1511.06281}{}{}{}.
\PrintBackRefs{\CurrentBib}

\bibitem [\protect \citeauthoryear {%
Banerjee%
, El~Ghaoui%
\BCBL {}\ \BBA {} d'Aspremont%
}{%
Banerjee%
\ \protect \BOthers {.}}{%
{\protect \APACyear {2008}}%
}]{%
banerjee2008model}
\APACinsertmetastar {%
banerjee2008model}%
\begin{APACrefauthors}%
Banerjee, O.%
, El~Ghaoui, L.%
\BCBL {}\ \BBA {} d'Aspremont, A.%
\end{APACrefauthors}%
\unskip\
\newblock
\APACrefYearMonthDay{2008}{}{}.
\newblock
{\BBOQ}\APACrefatitle {Model selection through sparse maximum likelihood
  estimation for multivariate {G}aussian or binary data} {Model selection
  through sparse maximum likelihood estimation for multivariate {G}aussian or
  binary data}.{\BBCQ}
\newblock
\APACjournalVolNumPages{Journal of Machine Learning Research}{9}{}{485--516}.
\PrintBackRefs{\CurrentBib}

\bibitem [\protect \citeauthoryear {%
Bartlett%
, Movellan%
\BCBL {}\ \BBA {} Sejnowski%
}{%
Bartlett%
\ \protect \BOthers {.}}{%
{\protect \APACyear {2002}}%
}]{%
bartlett2002face}
\APACinsertmetastar {%
bartlett2002face}%
\begin{APACrefauthors}%
Bartlett, M.%
, Movellan, J.%
\BCBL {}\ \BBA {} Sejnowski, T.%
\end{APACrefauthors}%
\unskip\
\newblock
\APACrefYearMonthDay{2002}{}{}.
\newblock
{\BBOQ}\APACrefatitle {Face recognition by independent component analysis}
  {Face recognition by independent component analysis}.{\BBCQ}
\newblock
\APACjournalVolNumPages{IEEE Transactions on Neural
  Networks}{13}{6}{1450--1464}.
\PrintBackRefs{\CurrentBib}

\bibitem [\protect \citeauthoryear {%
Bollob{\'a}s%
}{%
Bollob{\'a}s%
}{%
{\protect \APACyear {1998}}%
}]{%
bollobas1998modern}
\APACinsertmetastar {%
bollobas1998modern}%
\begin{APACrefauthors}%
Bollob{\'a}s, B.%
\end{APACrefauthors}%
\unskip\
\newblock
\APACrefYear{1998}.
\newblock
\APACrefbtitle {Modern graph theory} {Modern graph theory}\ (\BVOL~184).
\newblock
\APACaddressPublisher{}{Springer}.
\PrintBackRefs{\CurrentBib}

\bibitem [\protect \citeauthoryear {%
Campi%
, Parkkonen%
, Hari%
\BCBL {}\ \BBA {} Hyv\"arinen%
}{%
Campi%
\ \protect \BOthers {.}}{%
{\protect \APACyear {2013}}%
}]{%
Campi2013}
\APACinsertmetastar {%
Campi2013}%
\begin{APACrefauthors}%
Campi, C.%
, Parkkonen, L.%
, Hari, R.%
\BCBL {}\ \BBA {} Hyv\"arinen, A.%
\end{APACrefauthors}%
\unskip\
\newblock
\APACrefYearMonthDay{2013}{}{}.
\newblock
{\BBOQ}\APACrefatitle {Non-linear canonical correlation for joint analysis of
  {MEG} signals from two subjects} {Non-linear canonical correlation for joint
  analysis of {MEG} signals from two subjects}.{\BBCQ}
\newblock
\APACjournalVolNumPages{Frontiers in Neuroscience}{7}{107}{}.
\PrintBackRefs{\CurrentBib}

\bibitem [\protect \citeauthoryear {%
Cardoso%
}{%
Cardoso%
}{%
{\protect \APACyear {1998}}%
}]{%
cardoso1998multidimensional}
\APACinsertmetastar {%
cardoso1998multidimensional}%
\begin{APACrefauthors}%
Cardoso, J.%
\end{APACrefauthors}%
\unskip\
\newblock
\APACrefYearMonthDay{1998}{}{}.
\newblock
{\BBOQ}\APACrefatitle {Multidimensional independent component analysis}
  {Multidimensional independent component analysis}.{\BBCQ}
\newblock
\BIn{} \APACrefbtitle {Proceedings of the 1998 {IEEE} International Conference
  on Acoustics, Speech and Signal Processing, 1998} {Proceedings of the 1998
  {IEEE} international conference on acoustics, speech and signal processing,
  1998}\ (\BVOL~4, \BPGS\ 1941--1944).
\PrintBackRefs{\CurrentBib}

\bibitem [\protect \citeauthoryear {%
Coen-Cagli%
, Dayan%
\BCBL {}\ \BBA {} Schwartz%
}{%
Coen-Cagli%
\ \protect \BOthers {.}}{%
{\protect \APACyear {2012}}%
}]{%
coen2012cortical}
\APACinsertmetastar {%
coen2012cortical}%
\begin{APACrefauthors}%
Coen-Cagli, R.%
, Dayan, P.%
\BCBL {}\ \BBA {} Schwartz, O.%
\end{APACrefauthors}%
\unskip\
\newblock
\APACrefYearMonthDay{2012}{}{}.
\newblock
{\BBOQ}\APACrefatitle {Cortical Surround Interactions and Perceptual Salience
  via Natural Scene Statistics} {Cortical surround interactions and perceptual
  salience via natural scene statistics}.{\BBCQ}
\newblock
\APACjournalVolNumPages{{P}{L}o{S} Computational Biology}{8}{3}{e1002405}.
\PrintBackRefs{\CurrentBib}

\bibitem [\protect \citeauthoryear {%
Comon%
}{%
Comon%
}{%
{\protect \APACyear {1994}}%
}]{%
comon1994independent}
\APACinsertmetastar {%
comon1994independent}%
\begin{APACrefauthors}%
Comon, P.%
\end{APACrefauthors}%
\unskip\
\newblock
\APACrefYearMonthDay{1994}{}{}.
\newblock
{\BBOQ}\APACrefatitle {{Independent component analysis, a new concept?}}
  {{Independent component analysis, a new concept?}}{\BBCQ}
\newblock
\APACjournalVolNumPages{Signal Processing}{36}{3}{287--314}.
\PrintBackRefs{\CurrentBib}

\bibitem [\protect \citeauthoryear {%
Friedman%
, Hastie%
\BCBL {}\ \BBA {} Tibshirani%
}{%
Friedman%
\ \protect \BOthers {.}}{%
{\protect \APACyear {2008}}%
}]{%
friedman2008sparse}
\APACinsertmetastar {%
friedman2008sparse}%
\begin{APACrefauthors}%
Friedman, J.%
, Hastie, T.%
\BCBL {}\ \BBA {} Tibshirani, R.%
\end{APACrefauthors}%
\unskip\
\newblock
\APACrefYearMonthDay{2008}{}{}.
\newblock
{\BBOQ}\APACrefatitle {Sparse inverse covariance estimation with the graphical
  lasso} {Sparse inverse covariance estimation with the graphical
  lasso}.{\BBCQ}
\newblock
\APACjournalVolNumPages{Biostatistics}{9}{3}{432--441}.
\PrintBackRefs{\CurrentBib}

\bibitem [\protect \citeauthoryear {%
G{\'o}mez-Herrero%
, Atienza%
, Egiazarian%
\BCBL {}\ \BBA {} Cantero%
}{%
G{\'o}mez-Herrero%
\ \protect \BOthers {.}}{%
{\protect \APACyear {2008}}%
}]{%
gomez2008measuring}
\APACinsertmetastar {%
gomez2008measuring}%
\begin{APACrefauthors}%
G{\'o}mez-Herrero, G.%
, Atienza, M.%
, Egiazarian, K.%
\BCBL {}\ \BBA {} Cantero, J.%
\end{APACrefauthors}%
\unskip\
\newblock
\APACrefYearMonthDay{2008}{}{}.
\newblock
{\BBOQ}\APACrefatitle {Measuring directional coupling between {EEG} sources}
  {Measuring directional coupling between {EEG} sources}.{\BBCQ}
\newblock
\APACjournalVolNumPages{Neuroimage}{43}{3}{497--508}.
\PrintBackRefs{\CurrentBib}

\bibitem [\protect \citeauthoryear {%
Gutmann%
\ \BBA {} Hirayama%
}{%
Gutmann%
\ \BBA {} Hirayama%
}{%
{\protect \APACyear {2011}}%
}]{%
Gutmann2011b}
\APACinsertmetastar {%
Gutmann2011b}%
\begin{APACrefauthors}%
Gutmann, M.%
\BCBT {}\ \BBA {} Hirayama, J.%
\end{APACrefauthors}%
\unskip\
\newblock
\APACrefYearMonthDay{2011}{}{}.
\newblock
{\BBOQ}\APACrefatitle {{B}regman divergence as general framework to estimate
  unnormalized statistical models} {{B}regman divergence as general framework
  to estimate unnormalized statistical models}.{\BBCQ}
\newblock
\BIn{} \APACrefbtitle {Proc. Conf. on Uncertainty in Artificial Intelligence
  ({UAI})} {Proc. conf. on uncertainty in artificial intelligence ({UAI})}\
  (\BPG~283-290).
\newblock
\APACaddressPublisher{Corvallis, Oregon}{AUAI Press}.
\PrintBackRefs{\CurrentBib}

\bibitem [\protect \citeauthoryear {%
Gutmann%
\ \BBA {} Hyv{\"a}rinen%
}{%
Gutmann%
\ \BBA {} Hyv{\"a}rinen%
}{%
{\protect \APACyear {2011}}%
}]{%
Gutmann2011a}
\APACinsertmetastar {%
Gutmann2011a}%
\begin{APACrefauthors}%
Gutmann, M.%
\BCBT {}\ \BBA {} Hyv{\"a}rinen, A.%
\end{APACrefauthors}%
\unskip\
\newblock
\APACrefYearMonthDay{2011}{}{}.
\newblock
{\BBOQ}\APACrefatitle {{E}xtracting coactivated features from multiple data
  sets} {{E}xtracting coactivated features from multiple data sets}.{\BBCQ}
\newblock
\BIn{} \APACrefbtitle {Proc. Int. Conf. on Artificial Neural Networks
  ({ICANN})} {Proc. int. conf. on artificial neural networks ({ICANN})}\
  (\BVOL\ 6791, \BPGS\ 323--330).
\newblock
\APACaddressPublisher{Berlin, Heidelberg}{Springer}.
\PrintBackRefs{\CurrentBib}

\bibitem [\protect \citeauthoryear {%
Gutmann%
\ \BBA {} Hyv{\"a}rinen%
}{%
Gutmann%
\ \BBA {} Hyv{\"a}rinen%
}{%
{\protect \APACyear {2012}}%
}]{%
Gutmann2012a}
\APACinsertmetastar {%
Gutmann2012a}%
\begin{APACrefauthors}%
Gutmann, M.%
\BCBT {}\ \BBA {} Hyv{\"a}rinen, A.%
\end{APACrefauthors}%
\unskip\
\newblock
\APACrefYearMonthDay{2012}{}{}.
\newblock
{\BBOQ}\APACrefatitle {{N}oise-contrastive estimation of unnormalized
  statistical models, with applications to natural image statistics}
  {{N}oise-contrastive estimation of unnormalized statistical models, with
  applications to natural image statistics}.{\BBCQ}
\newblock
\APACjournalVolNumPages{Journal of Machine Learning Research}{13}{}{307--361}.
\PrintBackRefs{\CurrentBib}

\bibitem [\protect \citeauthoryear {%
Gutmann%
\ \BBA {} Hyv\"arinen%
}{%
Gutmann%
\ \BBA {} Hyv\"arinen%
}{%
{\protect \APACyear {2013}}%
{\protect \APACexlab {{\protect \BCnt {1}}}}}]{%
Gutmann2013b}
\APACinsertmetastar {%
Gutmann2013b}%
\begin{APACrefauthors}%
Gutmann, M.%
\BCBT {}\ \BBA {} Hyv\"arinen, A.%
\end{APACrefauthors}%
\unskip\
\newblock
\APACrefYearMonthDay{2013{\protect \BCnt {1}}}{}{}.
\newblock
{\BBOQ}\APACrefatitle {Estimation of unnormalized statistical models without
  numerical integration} {Estimation of unnormalized statistical models without
  numerical integration}.{\BBCQ}
\newblock
\BIn{} \APACrefbtitle {Proc Workshop on Information Theoretic Methods in
  Science and Engineering.} {Proc workshop on information theoretic methods in
  science and engineering.}
\PrintBackRefs{\CurrentBib}

\bibitem [\protect \citeauthoryear {%
Gutmann%
\ \BBA {} Hyv\"arinen%
}{%
Gutmann%
\ \BBA {} Hyv\"arinen%
}{%
{\protect \APACyear {2013}}%
{\protect \APACexlab {{\protect \BCnt {2}}}}}]{%
Gutmann2013}
\APACinsertmetastar {%
Gutmann2013}%
\begin{APACrefauthors}%
Gutmann, M.%
\BCBT {}\ \BBA {} Hyv\"arinen, A.%
\end{APACrefauthors}%
\unskip\
\newblock
\APACrefYearMonthDay{2013{\protect \BCnt {2}}}{}{}.
\newblock
{\BBOQ}\APACrefatitle {A three-layer model of natural image statistics} {A
  three-layer model of natural image statistics}.{\BBCQ}
\newblock
\APACjournalVolNumPages{Journal of Physiology-Paris}{107}{5}{369--398}.
\PrintBackRefs{\CurrentBib}

\bibitem [\protect \citeauthoryear {%
Gutmann%
, Laparra%
, Hyv\"arinen%
\BCBL {}\ \BBA {} Malo%
}{%
Gutmann%
\ \protect \BOthers {.}}{%
{\protect \APACyear {2014}}%
}]{%
Gutmann2014}
\APACinsertmetastar {%
Gutmann2014}%
\begin{APACrefauthors}%
Gutmann, M.%
, Laparra, V.%
, Hyv\"arinen, A.%
\BCBL {}\ \BBA {} Malo, J.%
\end{APACrefauthors}%
\unskip\
\newblock
\APACrefYearMonthDay{2014}{}{}.
\newblock
{\BBOQ}\APACrefatitle {Spatio-Chromatic Adaptation via Higher-Order Canonical
  Correlation Analysis of Natural Images} {Spatio-chromatic adaptation via
  higher-order canonical correlation analysis of natural images}.{\BBCQ}
\newblock
\APACjournalVolNumPages{PLOS ONE}{9}{2}{e86481--}.
\PrintBackRefs{\CurrentBib}

\bibitem [\protect \citeauthoryear {%
He%
, Qi%
, Kavukcuoglu%
\BCBL {}\ \BBA {} Park%
}{%
He%
\ \protect \BOthers {.}}{%
{\protect \APACyear {2012}}%
}]{%
NIPS2012_4636}
\APACinsertmetastar {%
NIPS2012_4636}%
\begin{APACrefauthors}%
He, Y.%
, Qi, Y.%
, Kavukcuoglu, K.%
\BCBL {}\ \BBA {} Park, H.%
\end{APACrefauthors}%
\unskip\
\newblock
\APACrefYearMonthDay{2012}{}{}.
\newblock
{\BBOQ}\APACrefatitle {Learning the Dependency Structure of Latent Factors}
  {Learning the dependency structure of latent factors}.{\BBCQ}
\newblock
\BIn{} \APACrefbtitle {Advances in Neural Information Processing Systems}
  {Advances in neural information processing systems}\ (\BPGS\ 2366--2374).
\PrintBackRefs{\CurrentBib}

\bibitem [\protect \citeauthoryear {%
Heeger%
}{%
Heeger%
}{%
{\protect \APACyear {1992}}%
}]{%
heeger1992normalization}
\APACinsertmetastar {%
heeger1992normalization}%
\begin{APACrefauthors}%
Heeger, D\BPBI J.%
\end{APACrefauthors}%
\unskip\
\newblock
\APACrefYearMonthDay{1992}{}{}.
\newblock
{\BBOQ}\APACrefatitle {Normalization of cell responses in cat striate cortex}
  {Normalization of cell responses in cat striate cortex}.{\BBCQ}
\newblock
\APACjournalVolNumPages{Visual neuroscience}{9}{2}{181--197}.
\PrintBackRefs{\CurrentBib}

\bibitem [\protect \citeauthoryear {%
Hinton%
}{%
Hinton%
}{%
{\protect \APACyear {2002}}%
}]{%
hinton2002training}
\APACinsertmetastar {%
hinton2002training}%
\begin{APACrefauthors}%
Hinton, G.%
\end{APACrefauthors}%
\unskip\
\newblock
\APACrefYearMonthDay{2002}{}{}.
\newblock
{\BBOQ}\APACrefatitle {Training products of experts by minimizing contrastive
  divergence} {Training products of experts by minimizing contrastive
  divergence}.{\BBCQ}
\newblock
\APACjournalVolNumPages{Neural Computation}{14}{8}{1771--1800}.
\PrintBackRefs{\CurrentBib}

\bibitem [\protect \citeauthoryear {%
Horn%
\ \BBA {} Johnson%
}{%
Horn%
\ \BBA {} Johnson%
}{%
{\protect \APACyear {1985}}%
}]{%
horn1985matrix}
\APACinsertmetastar {%
horn1985matrix}%
\begin{APACrefauthors}%
Horn, R.%
\BCBT {}\ \BBA {} Johnson, C.%
\end{APACrefauthors}%
\unskip\
\newblock
\APACrefYear{1985}.
\newblock
\APACrefbtitle {Matrix analysis} {Matrix analysis}.
\newblock
\APACaddressPublisher{}{{C}ambridge {U}niversity {P}ress}.
\PrintBackRefs{\CurrentBib}

\bibitem [\protect \citeauthoryear {%
Hoyer%
\ \BBA {} Hyv{\"a}rinen%
}{%
Hoyer%
\ \BBA {} Hyv{\"a}rinen%
}{%
{\protect \APACyear {2002}}%
}]{%
hoyer2002multi}
\APACinsertmetastar {%
hoyer2002multi}%
\begin{APACrefauthors}%
Hoyer, P.%
\BCBT {}\ \BBA {} Hyv{\"a}rinen, A.%
\end{APACrefauthors}%
\unskip\
\newblock
\APACrefYearMonthDay{2002}{}{}.
\newblock
{\BBOQ}\APACrefatitle {A multi-layer sparse coding network learns contour
  coding from natural images} {A multi-layer sparse coding network learns
  contour coding from natural images}.{\BBCQ}
\newblock
\APACjournalVolNumPages{Vision Research}{42}{12}{1593--1605}.
\PrintBackRefs{\CurrentBib}

\bibitem [\protect \citeauthoryear {%
Hurri%
\ \BBA {} Hyv{\"a}rinen%
}{%
Hurri%
\ \BBA {} Hyv{\"a}rinen%
}{%
{\protect \APACyear {2003}}%
}]{%
hurri2003temporal}
\APACinsertmetastar {%
hurri2003temporal}%
\begin{APACrefauthors}%
Hurri, J.%
\BCBT {}\ \BBA {} Hyv{\"a}rinen, A.%
\end{APACrefauthors}%
\unskip\
\newblock
\APACrefYearMonthDay{2003}{}{}.
\newblock
{\BBOQ}\APACrefatitle {Temporal and spatiotemporal coherence in simple-cell
  responses: a generative model of natural image sequences} {Temporal and
  spatiotemporal coherence in simple-cell responses: a generative model of
  natural image sequences}.{\BBCQ}
\newblock
\APACjournalVolNumPages{Network: Computation in Neural
  Systems}{14}{3}{527--551}.
\PrintBackRefs{\CurrentBib}

\bibitem [\protect \citeauthoryear {%
Hyv{\"a}rinen%
}{%
Hyv{\"a}rinen%
}{%
{\protect \APACyear {2005}}%
}]{%
hyvarinen2005estimation}
\APACinsertmetastar {%
hyvarinen2005estimation}%
\begin{APACrefauthors}%
Hyv{\"a}rinen, A.%
\end{APACrefauthors}%
\unskip\
\newblock
\APACrefYearMonthDay{2005}{}{}.
\newblock
{\BBOQ}\APACrefatitle {Estimation of non-normalized statistical models by score
  matching} {Estimation of non-normalized statistical models by score
  matching}.{\BBCQ}
\newblock
\APACjournalVolNumPages{Journal of Machine Learning Research}{6}{}{695--709}.
\PrintBackRefs{\CurrentBib}

\bibitem [\protect \citeauthoryear {%
Hyv{\"a}rinen%
}{%
Hyv{\"a}rinen%
}{%
{\protect \APACyear {2007}}%
}]{%
hyvarinen2007some}
\APACinsertmetastar {%
hyvarinen2007some}%
\begin{APACrefauthors}%
Hyv{\"a}rinen, A.%
\end{APACrefauthors}%
\unskip\
\newblock
\APACrefYearMonthDay{2007}{}{}.
\newblock
{\BBOQ}\APACrefatitle {Some extensions of score matching} {Some extensions of
  score matching}.{\BBCQ}
\newblock
\APACjournalVolNumPages{Computational Statistics \& Data
  Analysis}{51}{5}{2499--2512}.
\PrintBackRefs{\CurrentBib}

\bibitem [\protect \citeauthoryear {%
Hyv{\"a}rinen%
, Gutmann%
\BCBL {}\ \BBA {} Hoyer%
}{%
Hyv{\"a}rinen%
\ \protect \BOthers {.}}{%
{\protect \APACyear {2005}}%
}]{%
hyvarinen2005statistical}
\APACinsertmetastar {%
hyvarinen2005statistical}%
\begin{APACrefauthors}%
Hyv{\"a}rinen, A.%
, Gutmann, M.%
\BCBL {}\ \BBA {} Hoyer, P.%
\end{APACrefauthors}%
\unskip\
\newblock
\APACrefYearMonthDay{2005}{}{}.
\newblock
{\BBOQ}\APACrefatitle {Statistical model of natural stimuli predicts edge-like
  pooling of spatial frequency channels in {V2}} {Statistical model of natural
  stimuli predicts edge-like pooling of spatial frequency channels in
  {V2}}.{\BBCQ}
\newblock
\APACjournalVolNumPages{BMC Neuroscience}{6}{}{12}.
\PrintBackRefs{\CurrentBib}

\bibitem [\protect \citeauthoryear {%
Hyv{\"a}rinen%
\ \BBA {} Hoyer%
}{%
Hyv{\"a}rinen%
\ \BBA {} Hoyer%
}{%
{\protect \APACyear {2000}}%
}]{%
hyvarinen2000emergence}
\APACinsertmetastar {%
hyvarinen2000emergence}%
\begin{APACrefauthors}%
Hyv{\"a}rinen, A.%
\BCBT {}\ \BBA {} Hoyer, P.%
\end{APACrefauthors}%
\unskip\
\newblock
\APACrefYearMonthDay{2000}{}{}.
\newblock
{\BBOQ}\APACrefatitle {Emergence of phase- and shift-invariant features by
  decomposition of natural images into independent feature subspaces}
  {Emergence of phase- and shift-invariant features by decomposition of natural
  images into independent feature subspaces}.{\BBCQ}
\newblock
\APACjournalVolNumPages{Neural Computation}{12}{7}{1705--1720}.
\PrintBackRefs{\CurrentBib}

\bibitem [\protect \citeauthoryear {%
Hyv{\"a}rinen%
, Hoyer%
\BCBL {}\ \BBA {} Inki%
}{%
Hyv{\"a}rinen%
\ \protect \BOthers {.}}{%
{\protect \APACyear {2001}}%
}]{%
hyvarinen2001topographic}
\APACinsertmetastar {%
hyvarinen2001topographic}%
\begin{APACrefauthors}%
Hyv{\"a}rinen, A.%
, Hoyer, P.%
\BCBL {}\ \BBA {} Inki, M.%
\end{APACrefauthors}%
\unskip\
\newblock
\APACrefYearMonthDay{2001}{}{}.
\newblock
{\BBOQ}\APACrefatitle {{Topographic independent component analysis}}
  {{Topographic independent component analysis}}.{\BBCQ}
\newblock
\APACjournalVolNumPages{Neural Computation}{13}{7}{1527--1558}.
\PrintBackRefs{\CurrentBib}

\bibitem [\protect \citeauthoryear {%
Hyv{\"a}rinen%
, Hurri%
\BCBL {}\ \BBA {} Hoyer%
}{%
Hyv{\"a}rinen%
\ \protect \BOthers {.}}{%
{\protect \APACyear {2009}}%
}]{%
Hyvarinen2009}
\APACinsertmetastar {%
Hyvarinen2009}%
\begin{APACrefauthors}%
Hyv{\"a}rinen, A.%
, Hurri, J.%
\BCBL {}\ \BBA {} Hoyer, P.%
\end{APACrefauthors}%
\unskip\
\newblock
\APACrefYear{2009}.
\newblock
\APACrefbtitle {{N}atural {I}mage {S}tatistics} {{N}atural {I}mage
  {S}tatistics}.
\newblock
\APACaddressPublisher{}{Springer}.
\PrintBackRefs{\CurrentBib}

\bibitem [\protect \citeauthoryear {%
Hyv{\"a}rinen%
\ \BBA {} Oja%
}{%
Hyv{\"a}rinen%
\ \BBA {} Oja%
}{%
{\protect \APACyear {2000}}%
}]{%
hyvarinen2000independent}
\APACinsertmetastar {%
hyvarinen2000independent}%
\begin{APACrefauthors}%
Hyv{\"a}rinen, A.%
\BCBT {}\ \BBA {} Oja, E.%
\end{APACrefauthors}%
\unskip\
\newblock
\APACrefYearMonthDay{2000}{}{}.
\newblock
{\BBOQ}\APACrefatitle {{Independent component analysis: algorithms and
  applications}} {{Independent component analysis: algorithms and
  applications}}.{\BBCQ}
\newblock
\APACjournalVolNumPages{Neural Networks}{13}{4--5}{411--430}.
\PrintBackRefs{\CurrentBib}

\bibitem [\protect \citeauthoryear {%
{International~Phonetic~Association}%
}{%
{International~Phonetic~Association}%
}{%
{\protect \APACyear {1999}}%
}]{%
international1999handbook}
\APACinsertmetastar {%
international1999handbook}%
\begin{APACrefauthors}%
{International~Phonetic~Association}.%
\end{APACrefauthors}%
\unskip\
\newblock
\APACrefYear{1999}.
\newblock
\APACrefbtitle {Handbook of the International Phonetic Association: A guide to
  the use of the International Phonetic Alphabet} {Handbook of the
  international phonetic association: A guide to the use of the international
  phonetic alphabet}.
\newblock
\APACaddressPublisher{}{Cambridge University Press}.
\PrintBackRefs{\CurrentBib}

\bibitem [\protect \citeauthoryear {%
Karklin%
\ \BBA {} Lewicki%
}{%
Karklin%
\ \BBA {} Lewicki%
}{%
{\protect \APACyear {2005}}%
}]{%
karklin2005hierarchical}
\APACinsertmetastar {%
karklin2005hierarchical}%
\begin{APACrefauthors}%
Karklin, Y.%
\BCBT {}\ \BBA {} Lewicki, M.%
\end{APACrefauthors}%
\unskip\
\newblock
\APACrefYearMonthDay{2005}{}{}.
\newblock
{\BBOQ}\APACrefatitle {A hierarchical {B}ayesian model for learning nonlinear
  statistical regularities in nonstationary natural signals} {A hierarchical
  {B}ayesian model for learning nonlinear statistical regularities in
  nonstationary natural signals}.{\BBCQ}
\newblock
\APACjournalVolNumPages{Neural Computation}{17}{2}{397--423}.
\PrintBackRefs{\CurrentBib}

\bibitem [\protect \citeauthoryear {%
Klein%
, K{\"o}nig%
\BCBL {}\ \BBA {} K{\"o}rding%
}{%
Klein%
\ \protect \BOthers {.}}{%
{\protect \APACyear {2003}}%
}]{%
klein2003sparse}
\APACinsertmetastar {%
klein2003sparse}%
\begin{APACrefauthors}%
Klein, D\BPBI J.%
, K{\"o}nig, P.%
\BCBL {}\ \BBA {} K{\"o}rding, K\BPBI P.%
\end{APACrefauthors}%
\unskip\
\newblock
\APACrefYearMonthDay{2003}{}{}.
\newblock
{\BBOQ}\APACrefatitle {Sparse spectrotemporal coding of sounds} {Sparse
  spectrotemporal coding of sounds}.{\BBCQ}
\newblock
\APACjournalVolNumPages{EURASIP Journal on Applied Signal
  Processing}{2003}{}{659--667}.
\PrintBackRefs{\CurrentBib}

\bibitem [\protect \citeauthoryear {%
Kohonen%
}{%
Kohonen%
}{%
{\protect \APACyear {1995}}%
}]{%
kohonen1995self}
\APACinsertmetastar {%
kohonen1995self}%
\begin{APACrefauthors}%
Kohonen, T.%
\end{APACrefauthors}%
\unskip\
\newblock
\APACrefYear{1995}.
\newblock
\APACrefbtitle {Self-organizing maps} {Self-organizing maps}\ (\BVOL~30).
\newblock
\APACaddressPublisher{}{Springer-Verlag Berlin Heidelberg}.
\PrintBackRefs{\CurrentBib}

\bibitem [\protect \citeauthoryear {%
Kohonen%
}{%
Kohonen%
}{%
{\protect \APACyear {1996}}%
}]{%
kohonen1996emergence}
\APACinsertmetastar {%
kohonen1996emergence}%
\begin{APACrefauthors}%
Kohonen, T.%
\end{APACrefauthors}%
\unskip\
\newblock
\APACrefYearMonthDay{1996}{}{}.
\newblock
{\BBOQ}\APACrefatitle {Emergence of invariant-feature detectors in the
  adaptive-subspace self-organizing map} {Emergence of invariant-feature
  detectors in the adaptive-subspace self-organizing map}.{\BBCQ}
\newblock
\APACjournalVolNumPages{Biological cybernetics}{75}{4}{281--291}.
\PrintBackRefs{\CurrentBib}

\bibitem [\protect \citeauthoryear {%
K{\"o}ster%
\ \BBA {} Hyv{\"a}rinen%
}{%
K{\"o}ster%
\ \BBA {} Hyv{\"a}rinen%
}{%
{\protect \APACyear {2010}}%
}]{%
koster2010two}
\APACinsertmetastar {%
koster2010two}%
\begin{APACrefauthors}%
K{\"o}ster, U.%
\BCBT {}\ \BBA {} Hyv{\"a}rinen, A.%
\end{APACrefauthors}%
\unskip\
\newblock
\APACrefYearMonthDay{2010}{}{}.
\newblock
{\BBOQ}\APACrefatitle {A two-layer model of natural stimuli estimated with
  score matching} {A two-layer model of natural stimuli estimated with score
  matching}.{\BBCQ}
\newblock
\APACjournalVolNumPages{Neural Computation}{22}{9}{2308--2333}.
\PrintBackRefs{\CurrentBib}

\bibitem [\protect \citeauthoryear {%
Mairal%
, Jenatton%
, Obozinski%
\BCBL {}\ \BBA {} Bach%
}{%
Mairal%
\ \protect \BOthers {.}}{%
{\protect \APACyear {2011}}%
}]{%
mairal2011convex}
\APACinsertmetastar {%
mairal2011convex}%
\begin{APACrefauthors}%
Mairal, J.%
, Jenatton, R.%
, Obozinski, G.%
\BCBL {}\ \BBA {} Bach, F.%
\end{APACrefauthors}%
\unskip\
\newblock
\APACrefYearMonthDay{2011}{}{}.
\newblock
{\BBOQ}\APACrefatitle {Convex and Network Flow Optimization for Structured
  Sparsity} {Convex and network flow optimization for structured
  sparsity}.{\BBCQ}
\newblock
\APACjournalVolNumPages{Journal of Machine Learning
  Research}{12}{}{2681--2720}.
\PrintBackRefs{\CurrentBib}

\bibitem [\protect \citeauthoryear {%
Olshausen%
\ \BBA {} Field%
}{%
Olshausen%
\ \BBA {} Field%
}{%
{\protect \APACyear {1996}}%
}]{%
olshause1996emergence}
\APACinsertmetastar {%
olshause1996emergence}%
\begin{APACrefauthors}%
Olshausen, B.%
\BCBT {}\ \BBA {} Field, D.%
\end{APACrefauthors}%
\unskip\
\newblock
\APACrefYearMonthDay{1996}{}{}.
\newblock
{\BBOQ}\APACrefatitle {Emergence of simple-cell receptive field properties by
  learning a sparse code for natural images} {Emergence of simple-cell
  receptive field properties by learning a sparse code for natural
  images}.{\BBCQ}
\newblock
\APACjournalVolNumPages{Nature}{381}{}{607--609}.
\PrintBackRefs{\CurrentBib}

\bibitem [\protect \citeauthoryear {%
Osindero%
, Welling%
\BCBL {}\ \BBA {} Hinton%
}{%
Osindero%
\ \protect \BOthers {.}}{%
{\protect \APACyear {2006}}%
}]{%
osindero2006topographic}
\APACinsertmetastar {%
osindero2006topographic}%
\begin{APACrefauthors}%
Osindero, S.%
, Welling, M.%
\BCBL {}\ \BBA {} Hinton, G.%
\end{APACrefauthors}%
\unskip\
\newblock
\APACrefYearMonthDay{2006}{}{}.
\newblock
{\BBOQ}\APACrefatitle {Topographic product models applied to natural scene
  statistics} {Topographic product models applied to natural scene
  statistics}.{\BBCQ}
\newblock
\APACjournalVolNumPages{Neural Computation}{18}{2}{381--414}.
\PrintBackRefs{\CurrentBib}

\bibitem [\protect \citeauthoryear {%
Ostrowski%
}{%
Ostrowski%
}{%
{\protect \APACyear {1937}}%
}]{%
ostrowski1937sur}
\APACinsertmetastar {%
ostrowski1937sur}%
\begin{APACrefauthors}%
Ostrowski, A.%
\end{APACrefauthors}%
\unskip\
\newblock
\APACrefYearMonthDay{1937}{}{}.
\newblock
{\BBOQ}\APACrefatitle {Sur la d{\'e}termination des bornes inf{\'e}rieures pour
  une classe des d{\'e}terminants} {Sur la d{\'e}termination des bornes
  inf{\'e}rieures pour une classe des d{\'e}terminants}.{\BBCQ}
\newblock
\APACjournalVolNumPages{Bull. Sci. Math}{61}{2}{19--32}.
\PrintBackRefs{\CurrentBib}

\bibitem [\protect \citeauthoryear {%
Pihlaja%
, Gutmann%
\BCBL {}\ \BBA {} Hyv\"arinen%
}{%
Pihlaja%
\ \protect \BOthers {.}}{%
{\protect \APACyear {2010}}%
}]{%
Pihlaja2010}
\APACinsertmetastar {%
Pihlaja2010}%
\begin{APACrefauthors}%
Pihlaja, M.%
, Gutmann, M.%
\BCBL {}\ \BBA {} Hyv\"arinen, A.%
\end{APACrefauthors}%
\unskip\
\newblock
\APACrefYearMonthDay{2010}{}{}.
\newblock
{\BBOQ}\APACrefatitle {{A} {F}amily of {C}omputationally {E}fficient and
  {S}imple {E}stimators for {U}nnormalized {S}tatistical {M}odels} {{A}
  {F}amily of {C}omputationally {E}fficient and {S}imple {E}stimators for
  {U}nnormalized {S}tatistical {M}odels}.{\BBCQ}
\newblock
\BIn{} \APACrefbtitle {Proc. Conf. on Uncertainty in Artificial Intelligence
  ({UAI})} {Proc. conf. on uncertainty in artificial intelligence ({UAI})}\
  (\BPGS\ 442--449).
\newblock
\APACaddressPublisher{Corvallis, Oregon}{AUAI Press}.
\PrintBackRefs{\CurrentBib}

\bibitem [\protect \citeauthoryear {%
Sasaki%
, Gutmann%
, Shouno%
\BCBL {}\ \BBA {} Hyv{\"a}rinen%
}{%
Sasaki%
\ \protect \BOthers {.}}{%
{\protect \APACyear {2013}}%
}]{%
sasaki2013correlated}
\APACinsertmetastar {%
sasaki2013correlated}%
\begin{APACrefauthors}%
Sasaki, H.%
, Gutmann, M.%
, Shouno, H.%
\BCBL {}\ \BBA {} Hyv{\"a}rinen, A.%
\end{APACrefauthors}%
\unskip\
\newblock
\APACrefYearMonthDay{2013}{}{}.
\newblock
{\BBOQ}\APACrefatitle {Correlated topographic analysis: estimating an ordering
  of correlated components} {Correlated topographic analysis: estimating an
  ordering of correlated components}.{\BBCQ}
\newblock
\APACjournalVolNumPages{Machine Learning}{92}{2-3}{285--317}.
\PrintBackRefs{\CurrentBib}

\bibitem [\protect \citeauthoryear {%
Sasaki%
, Gutmann%
, Shouno%
\BCBL {}\ \BBA {} Hyv{\"a}rinen%
}{%
Sasaki%
\ \protect \BOthers {.}}{%
{\protect \APACyear {2014}}%
}]{%
Sasaki2014esti}
\APACinsertmetastar {%
Sasaki2014esti}%
\begin{APACrefauthors}%
Sasaki, H.%
, Gutmann, M.%
, Shouno, H.%
\BCBL {}\ \BBA {} Hyv{\"a}rinen, A.%
\end{APACrefauthors}%
\unskip\
\newblock
\APACrefYearMonthDay{2014}{}{}.
\newblock
{\BBOQ}\APACrefatitle {Estimating Dependency Structures for non-{G}Aussian
  Components with Linear and Energy Correlations} {Estimating dependency
  structures for non-{G}aussian components with linear and energy
  correlations}.{\BBCQ}
\newblock
\BIn{} \APACrefbtitle {Proceedings of the 17th International Conference on
  Artificial Intelligence and Statistics ({AISTATS}), {JMLR}: {W}\&{CP}}
  {Proceedings of the 17th international conference on artificial intelligence
  and statistics ({AISTATS}), {JMLR}: {W}\&{CP}}\ (\BVOL~33, \BPG~868-876).
\PrintBackRefs{\CurrentBib}

\bibitem [\protect \citeauthoryear {%
Schwartz%
\ \BBA {} Simoncelli%
}{%
Schwartz%
\ \BBA {} Simoncelli%
}{%
{\protect \APACyear {2001}}%
}]{%
schwartz2001natural}
\APACinsertmetastar {%
schwartz2001natural}%
\begin{APACrefauthors}%
Schwartz, O.%
\BCBT {}\ \BBA {} Simoncelli, E\BPBI P.%
\end{APACrefauthors}%
\unskip\
\newblock
\APACrefYearMonthDay{2001}{}{}.
\newblock
{\BBOQ}\APACrefatitle {Natural signal statistics and sensory gain control}
  {Natural signal statistics and sensory gain control}.{\BBCQ}
\newblock
\APACjournalVolNumPages{Nature neuroscience}{4}{8}{819--825}.
\PrintBackRefs{\CurrentBib}

\bibitem [\protect \citeauthoryear {%
Shimizu%
, Hoyer%
, Hyv{\"a}rinen%
\BCBL {}\ \BBA {} Kerminen%
}{%
Shimizu%
\ \protect \BOthers {.}}{%
{\protect \APACyear {2006}}%
}]{%
shimizu2006linear}
\APACinsertmetastar {%
shimizu2006linear}%
\begin{APACrefauthors}%
Shimizu, S.%
, Hoyer, P.%
, Hyv{\"a}rinen, A.%
\BCBL {}\ \BBA {} Kerminen, A.%
\end{APACrefauthors}%
\unskip\
\newblock
\APACrefYearMonthDay{2006}{}{}.
\newblock
{\BBOQ}\APACrefatitle {A Linear Non-{G}Aussian Acyclic Model for Causal
  Discovery} {A linear non-{G}aussian acyclic model for causal
  discovery}.{\BBCQ}
\newblock
\APACjournalVolNumPages{Journal of Machine Learning Research}{7}{}{2003--2030}.
\PrintBackRefs{\CurrentBib}

\bibitem [\protect \citeauthoryear {%
Simoncelli%
}{%
Simoncelli%
}{%
{\protect \APACyear {1999}}%
}]{%
simoncelli1999modeling}
\APACinsertmetastar {%
simoncelli1999modeling}%
\begin{APACrefauthors}%
Simoncelli, E.%
\end{APACrefauthors}%
\unskip\
\newblock
\APACrefYearMonthDay{1999}{}{}.
\newblock
{\BBOQ}\APACrefatitle {Modeling the joint statistics of images in the wavelet
  domain} {Modeling the joint statistics of images in the wavelet
  domain}.{\BBCQ}
\newblock
\BIn{} \APACrefbtitle {Proc SPIE, 44th Annual Meeting} {Proc spie, 44th annual
  meeting}\ (\BVOL\ 3813, \BPGS\ 188--195).
\PrintBackRefs{\CurrentBib}

\bibitem [\protect \citeauthoryear {%
Terashima%
\ \BBA {} Hosoya%
}{%
Terashima%
\ \BBA {} Hosoya%
}{%
{\protect \APACyear {2009}}%
}]{%
terashima2009sparse}
\APACinsertmetastar {%
terashima2009sparse}%
\begin{APACrefauthors}%
Terashima, H.%
\BCBT {}\ \BBA {} Hosoya, H.%
\end{APACrefauthors}%
\unskip\
\newblock
\APACrefYearMonthDay{2009}{}{}.
\newblock
{\BBOQ}\APACrefatitle {Sparse codes of harmonic natural sounds and their
  modulatory interactions} {Sparse codes of harmonic natural sounds and their
  modulatory interactions}.{\BBCQ}
\newblock
\APACjournalVolNumPages{Network: Computation in Neural
  Systems}{20}{4}{253--267}.
\PrintBackRefs{\CurrentBib}

\bibitem [\protect \citeauthoryear {%
Terashima%
, Hosoya%
, Tani%
, Ichinohe%
\BCBL {}\ \BBA {} Okada%
}{%
Terashima%
\ \protect \BOthers {.}}{%
{\protect \APACyear {2013}}%
}]{%
terashima2013sparse}
\APACinsertmetastar {%
terashima2013sparse}%
\begin{APACrefauthors}%
Terashima, H.%
, Hosoya, H.%
, Tani, T.%
, Ichinohe, N.%
\BCBL {}\ \BBA {} Okada, M.%
\end{APACrefauthors}%
\unskip\
\newblock
\APACrefYearMonthDay{2013}{}{}.
\newblock
{\BBOQ}\APACrefatitle {Sparse coding of harmonic vocalization in monkey
  auditory cortex} {Sparse coding of harmonic vocalization in monkey auditory
  cortex}.{\BBCQ}
\newblock
\APACjournalVolNumPages{Neurocomputing}{103}{}{14--21}.
\PrintBackRefs{\CurrentBib}

\bibitem [\protect \citeauthoryear {%
Terashima%
\ \BBA {} Okada%
}{%
Terashima%
\ \BBA {} Okada%
}{%
{\protect \APACyear {2012}}%
}]{%
terashima2012topo}
\APACinsertmetastar {%
terashima2012topo}%
\begin{APACrefauthors}%
Terashima, H.%
\BCBT {}\ \BBA {} Okada, M.%
\end{APACrefauthors}%
\unskip\
\newblock
\APACrefYearMonthDay{2012}{}{}.
\newblock
{\BBOQ}\APACrefatitle {The topographic unsupervised learning of natural sounds
  in the auditory cortex} {The topographic unsupervised learning of natural
  sounds in the auditory cortex}.{\BBCQ}
\newblock
\BIn{} \APACrefbtitle {Advances in Neural Information Processing Systems}
  {Advances in neural information processing systems}\ (\BPGS\ 2321--2329).
\PrintBackRefs{\CurrentBib}

\bibitem [\protect \citeauthoryear {%
Theis%
}{%
Theis%
}{%
{\protect \APACyear {2005}}%
}]{%
theis2005blind}
\APACinsertmetastar {%
theis2005blind}%
\begin{APACrefauthors}%
Theis, F.%
\end{APACrefauthors}%
\unskip\
\newblock
\APACrefYearMonthDay{2005}{}{}.
\newblock
{\BBOQ}\APACrefatitle {Blind signal separation into groups of dependent signals
  using joint block diagonalization} {Blind signal separation into groups of
  dependent signals using joint block diagonalization}.{\BBCQ}
\newblock
\BIn{} \APACrefbtitle {{IEEE} International Symposium on Circuits and Systems,
  2005} {{IEEE} international symposium on circuits and systems, 2005}\ (\BPGS\
  5878--5881).
\PrintBackRefs{\CurrentBib}

\bibitem [\protect \citeauthoryear {%
Vig{\'a}rio%
, S{\"a}rel{\"a}%
, Jousm{\"a}ki%
, H{\"a}m{\"a}l{\"a}inen%
\BCBL {}\ \BBA {} Oja%
}{%
Vig{\'a}rio%
\ \protect \BOthers {.}}{%
{\protect \APACyear {2000}}%
}]{%
vigario2000independent}
\APACinsertmetastar {%
vigario2000independent}%
\begin{APACrefauthors}%
Vig{\'a}rio, R.%
, S{\"a}rel{\"a}, J.%
, Jousm{\"a}ki, V.%
, H{\"a}m{\"a}l{\"a}inen, M.%
\BCBL {}\ \BBA {} Oja, E.%
\end{APACrefauthors}%
\unskip\
\newblock
\APACrefYearMonthDay{2000}{}{}.
\newblock
{\BBOQ}\APACrefatitle {Independent component approach to the analysis of {EEG}
  and {MEG} recordings} {Independent component approach to the analysis of
  {EEG} and {MEG} recordings}.{\BBCQ}
\newblock
\APACjournalVolNumPages{IEEE Transactions on Biomedical
  Engineering}{47}{5}{589--593}.
\PrintBackRefs{\CurrentBib}

\bibitem [\protect \citeauthoryear {%
Zoran%
\ \BBA {} Weiss%
}{%
Zoran%
\ \BBA {} Weiss%
}{%
{\protect \APACyear {2009}}%
}]{%
Zoran09}
\APACinsertmetastar {%
Zoran09}%
\begin{APACrefauthors}%
Zoran, D.%
\BCBT {}\ \BBA {} Weiss, Y.%
\end{APACrefauthors}%
\unskip\
\newblock
\APACrefYearMonthDay{2009}{}{}.
\newblock
{\BBOQ}\APACrefatitle {The {``Tree-Dependent Components"} of Natural Images are
  Edge Filters} {The {``Tree-Dependent Components"} of natural images are edge
  filters}.{\BBCQ}
\newblock
\BIn{} \APACrefbtitle {Advances in Neural Information Processing Systems}
  {Advances in neural information processing systems}\ (\BVOL~22, \BPGS\
  2340--2348).
\PrintBackRefs{\CurrentBib}

\end{thebibliography}
\end{document}